\documentclass[journal]{IEEEtran}

\usepackage[colorlinks=true,linkcolor=red,citecolor=blue,urlcolor=blue,]{hyperref}
\usepackage{graphicx}
\usepackage[cmex10]{amsmath}
\usepackage{amssymb}
\usepackage{tabularray}
\usepackage{ninecolors}
\usepackage{color}
\usepackage{booktabs}
\UseTblrLibrary{booktabs}
\UseTblrLibrary{diagbox}
\usepackage{multirow}
\usepackage{url}
\usepackage[style=ieee,backend=biber,hyperref=true]{biblatex}
\DeclareFieldFormat{bibliography}{\setlength{\baselineskip}{1.0\baselineskip}#1}

\newcommand\redbf[1]{\textcolor{red}{\textbf{#1}}}
\newcommand\bluebf[1]{\textcolor{blue}{\textbf{#1}}}

\makeatletter
\let\MYcaption\@makecaption
\makeatother
\usepackage[font=footnotesize]{subcaption}
\makeatletter
\let\@makecaption\MYcaption
\makeatother

\begin{document}
\title{PETDet: Proposal Enhancement for \\ Two-Stage Fine-Grained Object Detection}

\author{Wentao~Li,
  Danpei~Zhao,~\IEEEmembership{Member,~IEEE,}
  Bo Yuan,
  Yue Gao,
  and~Zhenwei~Shi,~\IEEEmembership{Senior Member,~IEEE}
  \thanks{This work was supported by the National Natural Science Foundation of China under Grant 62271018. (Corresponding
    author: Danpei Zhao.)}
  \thanks{Wentao Li, Danpei Zhao, Bo Yuan, and Zhenwei Shi are with the Image Processing Center, School of Astronautics, Beihang University, Beijing 100191, China (e-mail: canoe@buaa.edu.cn; zhaodanpei@buaa.edu.cn; yuanbobuaa@buaa.edu.cn; shizhenwei@buaa.edu.cn).}
  \thanks{Yue Gao is with the Image Processing Center, School of Astronautics, Beihang University, Beijing 100191, China,  and also with Space Star Technology Company Ltd., Beijing 100094, China (e-mail: bjlguniversity@163.com).}
}

\markboth{IEEE TRANSACTIONS ON GEOSCIENCE AND REMOTE SENSING}%
{Shell \MakeLowercase{\textit{et al.}}: Bare Demo of IEEEtran.cls for Journals}

\maketitle

\begin{abstract}
  Fine-grained object detection (FGOD) extends object detection with the capability of fine-grained recognition. In recent two-stage FGOD methods, the region proposal serves as a crucial link between detection and fine-grained recognition. However, current methods overlook that some proposal-related procedures inherited from general detection are not equally suitable for FGOD, limiting the multi-task learning from generation, representation, to utilization. In this paper, we present PETDet (Proposal Enhancement for Two-stage fine-grained object detection) to better handle the sub-tasks in two-stage FGOD methods. Firstly, an anchor-free Quality Oriented Proposal Network (QOPN) is proposed with dynamic label assignment and attention-based decomposition to generate high-quality oriented proposals. Additionally, we present a Bilinear Channel Fusion Network (BCFN) to extract independent and discriminative features of the proposals. Furthermore, we design a novel Adaptive Recognition Loss (ARL) which offers guidance for the R-CNN head to focus on high-quality proposals. Extensive experiments validate the effectiveness of PETDet. Quantitative analysis reveals that PETDet with ResNet50 reaches state-of-the-art performance on various FGOD datasets, including FAIR1M-v1.0 (42.96\,AP), FAIR1M-v2.0 (48.81\,AP), MAR20 (85.91\,AP) and ShipRSImageNet (74.90\,AP). The proposed method also achieves superior compatibility between accuracy and inference speed. Our code and models will be released at \textit{\url{https://github.com/canoe-Z/PETDet}}.
\end{abstract}

\IEEEpeerreviewmaketitle

\section{Introduction}
\begin{figure}[!t]
  \begin{center}
    \includegraphics[width=0.4\textwidth]{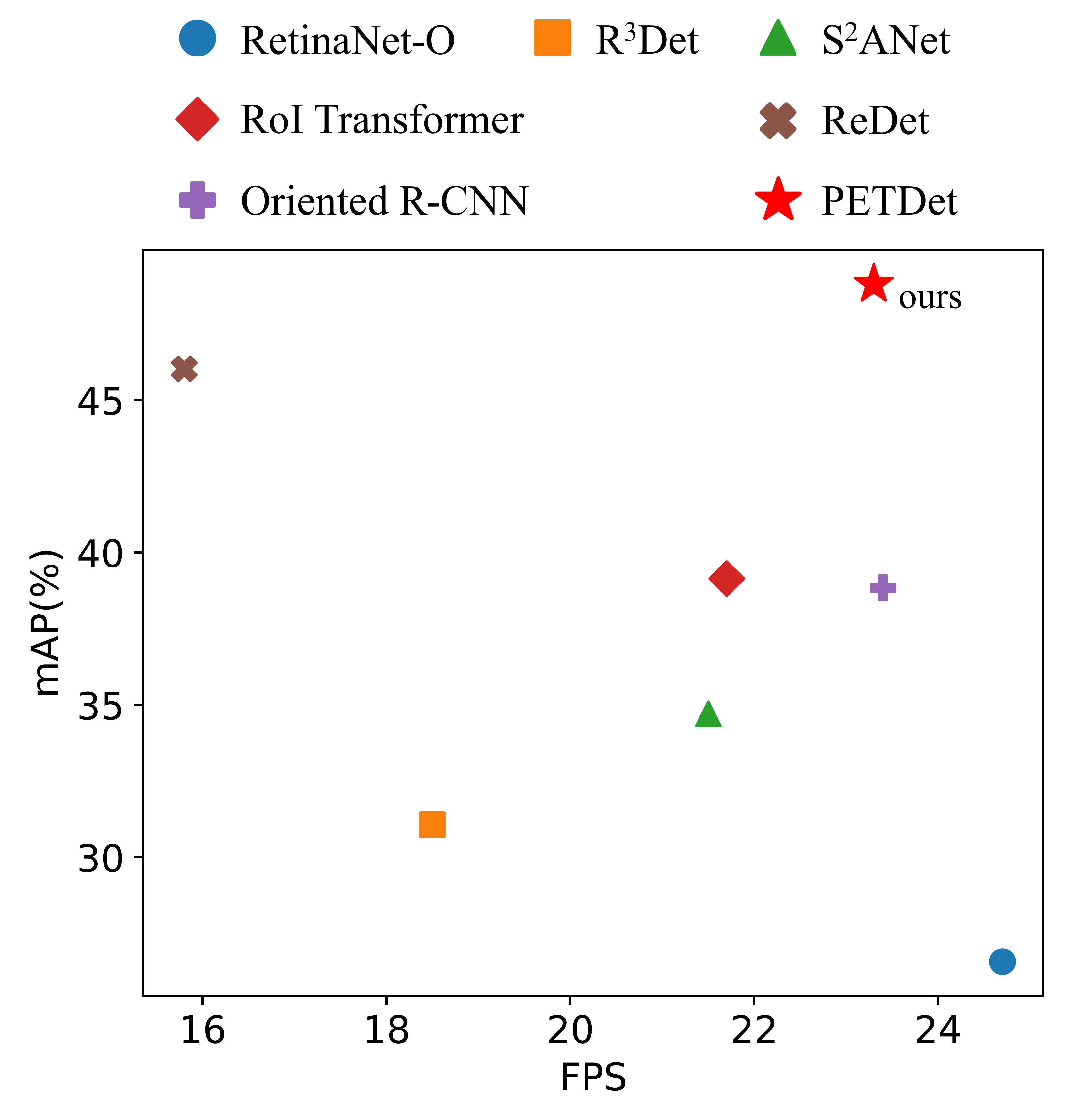}
  \end{center}
  \caption{Speed versus Accuracy on FAIR1M-2.0. Compared with other oriented object detectors, our PETDet can achieve state-of-the-art performance with competitive speed.}
  \label{fig:sva}
\end{figure}

\IEEEPARstart{F}INE-GRAINED object detection (FGOD) aims to accurately recognize fine-grained sub-categories while locating them simultaneously. For example, an effective fine-grained detector should not only correctly detect objects within the coarse category \textit{Airplane} but also recognize the fine-grained subcategory such as \textit{Airbus 350} or \textit{Boeing 747}. FGOD in aerial images has a broad application prospect, such as earth observation \cite{fair1m}, urban monitoring \cite{ubc}, and disaster control \cite{fdd}. However, compared with general object detection, FGOD presents a greater challenge due to the semantic confusion between classes caused by the large intra-class variance and the ambiguous inter-class differences.

\par
With the rapid development of high-resolution remote sensing technology, despite object detection has extensive successful applications in remote sensing \cite{dota}, \cite{low}, \cite{traffic}, it no longer meets the new demand for fine-grained recognition. In recent times, FGOD in aerial images has garnered widespread attention from the research community \cite{fgfe}, \cite{kd}, \cite{sfr}, \cite{osfd}, \cite{dosr}, \cite{sis}, \cite{acd}, \cite{iscl}, \cite{gicnet}. FGOD is a multi-task learning problem comprising foreground and background (FG/BG) classification, box regression and fine-grained recognition. Recent methods usually adopt the two-stage pipeline for better task decomposition via sparse region proposals. Besides, objects in aerial images often have varying orientations and aspect ratios, which cannot be adequately represented under the horizontal scheme. Current FGOD methods commonly perform oriented detection since oriented bounding boxes can facilitate precise fine-grained recognition with less redundant background regions. Based on two-stage pipeline and oriented scheme, many works focus on improving the fine-grained recognition performance with a variety of attention mechanisms or metric learning methods proposed. For examples, Zhou et al. \cite{fgfe} introduce attention-based group feature enhancement and sub-saliency feature learning. Cheng et al. \cite{sfr} propose a spatial and channel transformer to capture discriminative features and adopt deep metric learning to enhance the separability of fine-grained classes.

\par
Despite the notable progress, there are still significant limitations that need to be addressed. From the perspective of multi-task learning, properly handling the relationship between sub-tasks is critical in FGOD. For two-stage methods, the region proposal is the bridge linking the detection and recognition sub-tasks. However, the importance of proposals is neglected in previous work. In our study, we do not aim to develop new components to directly enhance fine-grained recognition. Instead, we try to unlock the potential of two-stage FGOD detectors by enhancing region proposals to improve collaborative optimization. We argue that some procedures related to region proposals, designed for general object detection may not be equally suitable for FGOD, which results in performance limitations. Specifically, the following three main issues hinder the performance of two-stage FGOD methods:

\par
\noindent
\textbf{(1) Generation:} For two-stage FGOD methods, generating high-quality proposals is one of the most critical task. High-quality classification results in fewer false-positive proposals, which allows the R-CNN head to focus more attention on recognition than FG/BG classification. With high-quality localization, the RoI features will align better with less redundant background. In contrast, a proposal with inaccurate regression which fails to fully enclose the target, will lead to the absence of essential discriminative features.

\par
\noindent
\textbf{(2) Representation}: In two-stage FGOD detectors, the first stage is responsible for FG/BG classification and proposal localization, while the second stage handles fine-grained recognition and bounding box refinement. However, features for both stages are extracted from the feature pyramid network (FPN) \cite{fpn} without being decoupled, which leads to confusion among tasks. Moreover, proposal representation based on single-level features is insufficient to support accurate fine-grained recognition in the second stage \cite{dosr}, \cite{sis}.

\par
\noindent
\textbf{(3) Utilization}: In previous two-stage methods, the R-CNN head takes proposals generated by the vanilla RPN as input, which include numerous false positives. Therefore, RoIs need to be sampled by a handcrafted positive/negative ratio to reduce the imbalance. Inheriting that procedure, the second stage in current two-stage FGOD methods still pays much attention to the FG/BG classification task. Even though the proposal quality had been enhanced, high-quality positive samples would not be fully utilized, which considerably hurts the learning of fine-grained recognition.

\par
In this paper, we focus the proposal enhancement and propose a novel two-stage FGOD method called PETDet (Proposal Enhancement for Two-stage fine-grained object detection). Our PETDet comprises three main components, each of which addresses one of above issues associated with region proposals. To improve the quality of proposals, a Quality Oriented Proposal Network (QOPN) is introduced, which is an anchor-free oriented proposal network with dynamic label assignment and attention-based decomposition. QOPN generates high-quality proposals to facilitate subsequent optimization with slight increase in computational cost. Additionally, inspired by the low-rank bilinear pooling \cite{mlb}, we present a Bilinear Channel Fusion Network (BCFN) to produce independent and discriminative features by cross-level fusion. To further enhance proposal utilization, we design an Adaptive Recognition Loss (ARL) for the R-CNN head. ARL jointly assesses the quality of each proposal based on the classification score and the refined IoU, up-weighting the loss assigned to high-quality samples. Since ARL can guide the R-CNN head to focus on certain samples, some inappropriate procedures such as the random sampling and the non-maximum suppression of proposals are discarded to maximize sample utilization.

\par
Extensive experimental results substantiate the effectiveness of proposed methods. Specifically, PETDet with ResNet-50-FPN achieves 48.81\,AP on FAIR1M-v2.0 and largely surpasses the strong baseline Oriented R-CNN by 4.91$\%$, with a negligible increase in inference time. Our method can also reach state-of-the-art performance on single-class FGOD dataset including MAR20 and ShipRSImageNet. As depicted in Fig. \ref{fig:sva}, compared to the baseline, PETDet has only a negligible impact on the inference speed. It can be concluded that our PETDet performs well in terms of both accuracy and efficiency.

\par
The main contributions of this work can be summarized as follows:
\begin{enumerate}
  \item We explore two-stage FGOD methods from a fresh perspective. By exploring three universal proposal-related bottlenecks which restrict multi-task learning for FGOD, we propose a novel FGOD framework with proposal enhancement strategy, providing guidance to improve existing two-stage methods.
  \item We propose an end-to-end FGOD method called PETDet, in which three novel synergistic modules are designed to resolve the inherent contradiction between object detection and fine-grained recognition sub-tasks within FGOD.
  \item The proposed PETDet sets new records on multiple FGOD datasets, including FAIR1M-V1.0, FAIR1M-V2.0, MAR20, and ShipRSImageNet.
\end{enumerate}
\par
The rest of the paper is organized as follows. Section \ref{sec:related} introduces a concise overview of the research progress related to our work. Section \ref{sec:method} introduces the details of proposed PETDet. Section \ref{sec:result} provides experiments on the effectiveness of the method. Finally, the conclusion is made in Section \ref{sec:conclusion}.

\section{Related Works} \label{sec:related}
In this section, we review the previous research on two-stage and one-stage methods in general object detection first, to clarify the relationship with RPN and one-stage detector. Then, as FGOD in aerial images commonly adopts oriented bounding boxes, we review related work on oriented object detection. Finally, we discuss the recent developments in the FGOD field.

\subsection{General Object Detection}
With the advance of deep learning, object detection has made significant progress. Convolution-based object detectors can be categorized into two-stage or one-stage methods. Two-stage methods adopt the region proposal network (RPN) to generate potential proposals, then perform RoI-wise box regression and classification in the second stage \cite{faster-rcnn}, \cite{cascade-rcnn}, \cite{libra-rcnn}, \cite{doublehead}, \cite{dynamic}. Due to the complex pipeline with numerous hand-designed components, two-stage methods are no longer the focus of community in general object detection more recently. However, for FGOD tasks, the two-stage pipeline has a unique advantage in task decomposition.

\par
In contrast, one-stage detectors detect objects directly without proposals. To address the problem of foreground-background class imbalance, Focal Loss is introduced \cite{retinanet} to down-weight well-classified samples. To further simplify the pipeline, anchor-free one-stage detectors use anchor-points \cite{fcos}, \cite{foveabox} or key-points \cite{cornernet}, \cite{centernet} instead of setting handcrafted anchors. Meanwhile, many advanced label assignment strategies \cite{atss}, \cite{paa}, \cite{ota} proposed to dynamically choose positive and negative samples. Moreover, kinds of soft label assignment methods \cite{gfl}, \cite{vfnet}, \cite{tood}, \cite{dw} have been proposed to alleviate the inconsistency of classification and localization.

\par
For two-stage FGOD methods, we argue that the quality of proposals is critical. RPN makes dense predictions like the one-stage detector, but the relatively simple architecture leads to low-quality predictions. Applying contrastive learning to proposals is a potential way to improve the true positive rate \cite{unsupervised}. However, it can not enhance localization quality concurrently. In our study, inspired by the probabilistic interpretation of two-stage detection \cite{centernet2}, we adopt advanced designs from modern one-stage detectors to generate high-quality proposals.

\subsection{Oriented Object Detection}
Comparing with general object detection, oriented object detection extends detectors with extra angle prediction. Aerial images is the most popular application scenario of oriented detector, where objects are usually arbitrary-oriented. Convolution-based oriented detectors can also be categorized into two-stage or one-stage. Two-stage methods perform oriented detection based on region proposals \cite{r2cnn}, \cite{rrpn}, \cite{GV}, \cite{roitrans}, \cite{orcnn}, \cite{redet}, \cite{aopg}, \cite{dodet}. RRPN \cite{rrpn} sets rotated anchors to generate oriented proposals. Gliding Vertex \cite{GV} glide the vertex of the horizontal bounding box to accurately describe an oriented object. Oriented R-CNN \cite{orcnn} directly learns oriented proposals from horizontal anchors with a six parameters representation.

\par
In recent years, one-stage oriented detectors have also made impressive progress \cite{scrdet}, \cite{r3det}, \cite{dal}, \cite{s2anet}, \cite{dal}, \cite{oreppoints}, \cite{cfc}, \cite{ridet}, \cite{tioe}. Some works pay attention on the feature misalignment problem. S2ANet \cite{s2anet} applies an anchor refinement network to generate oriented anchors. R3Det \cite{r3det} refines oriented bounding boxes by learning aligned feature maps. Some works concentrate on the representation of oriented bounding boxes to address the boundary problem. GWD \cite{gwd} and KLD \cite{kld} respectively adopted Gaussian Wallenstein distance and Kullback-Leibler divergence to measure the distance between boxes. PSC \cite{psc} predicts the orientation by mapping rotational periodicity of different cycles into phase of different frequencies. Apart from two-stage or one-stage convolution-based methods, recently several transformer-based oriented detectors have been proposed to make end-to-end detection. \cite{orienteddetr}, \cite{ao2}, \cite{ars}, \cite{d2q}, \cite{rhino}.

\par
Oriented bounding boxes encapsulate less redundant regions when representing arbitrary-oriented objects. This brings great benefits for FGOD in aerial imagery, where minimizing background redundancy can enhance the recognition task. Consequently, oriented prediction has become a popular choice in FGOD methods. In our study, we adopt the strong two-stage oriented detector Oriented R-CNN \cite{orcnn} as our baseline method.

\subsection{Fine-grained Object Detection}
Based on the development of general object detection and oriented object detection, FGOD in remote sensing has received increasing attention more recently \cite{fgfe}, \cite{kd}, \cite{sfr}, \cite{osfd}, \cite{dosr}, \cite{sis}, \cite{acd}, \cite{iscl}, \cite{gicnet}. Comparing with previous fine-grained recognition work based on classification task \cite{part}, \cite{racnn}, \cite{macnn}, \cite{mcloss}, FGOD requires simultaneous localization and fine-grained recognition. Current FGOD methods works mainly make efforts to alleviate the semantic confusion between fine-grained classes. Zhou et al. \cite{fgfe} propose attention-based group feature enhancement and sub-saliency feature learning. Wang et al.\cite{kd} introduce an extra backbone to learn fine-grained classification and adopt knowledge distillation to keep it lightweight. Ouyang et al. \cite{pcl} propose PCLDet to maximize the interclass distance and minimize the intraclass distance by prototypical contrastive learning. Cheng et al. \cite{sfr} propose SFRNet with a spatial and channel transformer to capture discriminative features and adopt metric learning to enhance the separability of fine-grained classes. Besides, several approaches are designed to perform FGOD on specific course-grained class. As for fine-grained ship detection, Ouyang et al. \cite{osfd} build MGANet with a self-attention network to exploit the global and local features. For fine-grained aircraft detection, Zeng et al. \cite{iscl} proposal ISCL to extract various discriminative feature by instance switching-based contrastive learnings.

\par
Most recent methods are designed following the two-stage paradigm to better decompose the sub-tasks of FGOD. However, their performance may still be impacted by some inappropriate designs of region proposals, including proposal generation, representation, and utilization, which are overlooked in previous studies. In our study, instead of directly enhancing fine-grained recognition, we focus on proposal enhancement to overcome these obstacles caused by region proposals.

\begin{figure*}[!t]
  \begin{center}
    \includegraphics[width=1.0\textwidth]{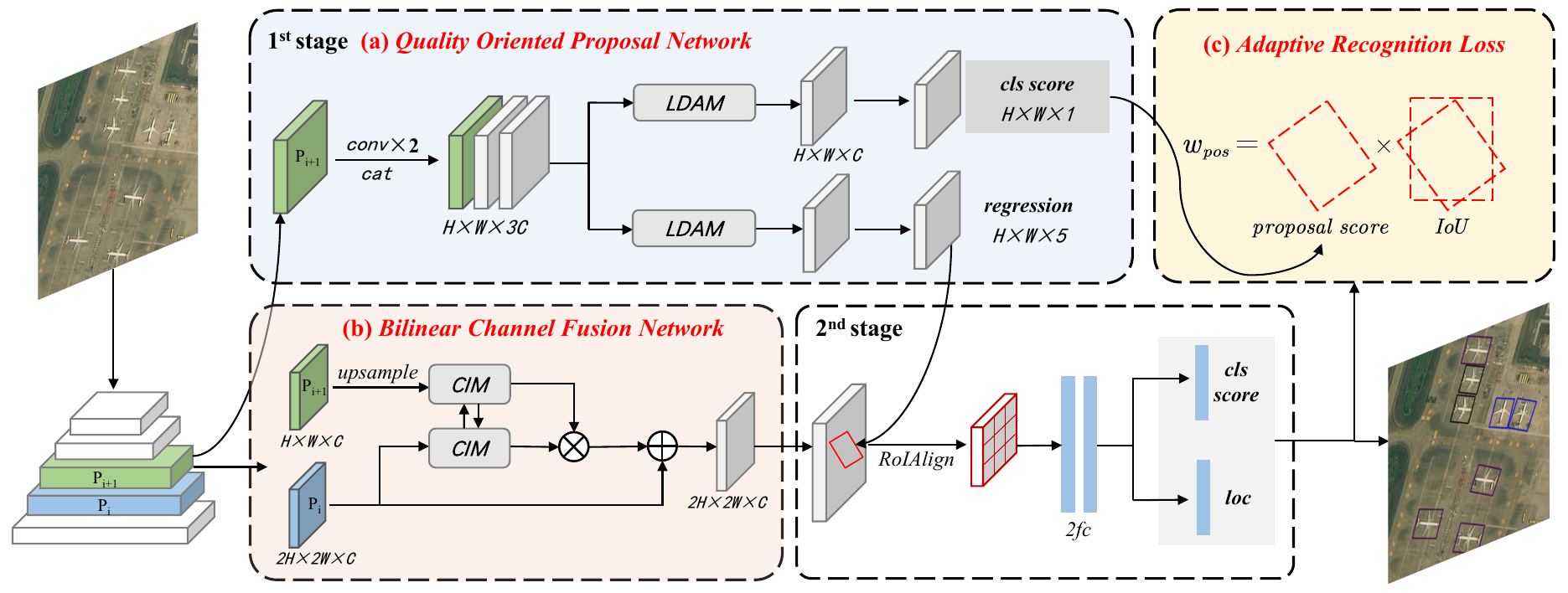}
  \end{center}
  \caption{The framework of our proposed PETDet model. PETDet is a two-stage detector designed for FGOD tasks by proposal enhancement, including: (a): Quality Oriented Proposal Network: generate high-quality oriented proposals to free the second stage from FG/BG classification and imprecise RoIs. (b): Bilinear Channel Fusion Network: fuse cross-level features bilinearly to enhance the representation of proposals. (c): Adaptive Recognition Loss: up-weight high-quality proposals facilitate fine-grained recognition learning. Note that QOPN takes single-level feature map $P_i$ from FPN as input while the second stage perform on cross-level feature map fused by BCFN.}
  \label{fig:petdet}
\end{figure*}

\section{Method} \label{sec:method}
In this section, we present a two-stage FGOD method named PETDet. Fig. \ref{fig:petdet} depicts the overall framework of the proposed method. As it shown, PETDet consists three proposed components for proposals enhancement. We will detail the Quality Oriented Proposal Network in Section \ref{sec:qopn}. The Bilinear Channel Fusion Network and Adaptive Recognition Loss are presented in Section \ref{sec:bcfn} and Section \ref{sec:arl}, respectively.

\subsection{Quality Oriented Proposal Network} \label{sec:qopn}
We propose QOPN to generate well-localization proposals with fewer false positives introduced. The idea is inspired by CenterNet2 \cite{centernet2}, in which a single-stage object detector replaces the vanilla RPN for high-quality proposal generation. However, setting more parameters in the first stage will cause the imbalance between stages and be harmful for collaborative optimization in FGOD tasks. On the contrary, advanced training strategies of one-stage detectors instead of the direct replacement are adopted in QOPN. Specifically, QOPN adopt anchor-free paradigm, learning the offsets and angle based on prior points like FCOS \cite{fcos} instead of setting horizontal anchors to avoid the problem that horizontal anchors may be hard to pair with oriented ground truths. After that, QOPN applies the adaptive training sample selection (ATSS) \cite{atss} for dynamic label assignment rather than setting a fixed threshold, enabling adaptive sample selection with fewer hyperparameters. With those improvements, QOPN can generate high-quality oriented proposals for less background redundancy to facilitate fine-grained recognition without extra trainable parameters.

\par
To further improve the quality of proposals, we also enhance the network architectures of QOPN. Instead of introducing decoupled branches for localization and classification, which is widely used in one-stage detectors, only several extra shared convolution layers are set in QOPN to avoid a marked increase in computations. Let $X^{fpn}\in \mathbb{R} ^{H\times W\times C}$ signifies the FPN features, where $H$, $W$ and $C$ denote the height, width, and the number of channels of the feature maps, respectively. We apply consecutive conv layers shared for both localization and classification to extract multiscale features $X^{conv}=\left\{ X^1,X^2,...,X^k \right\} ,k\in \left\{ 1,2,...,N \right\}$, where $N$ denotes the number of $3\times3$ conv layers. In our implementation, $N=2$ is our default setting since experiments prove that two shared conv layers is adequate with LDAM detailed in the following paragraphs.

\begin{figure}[!t]
  \begin{center}
    \includegraphics[width=0.45\textwidth]{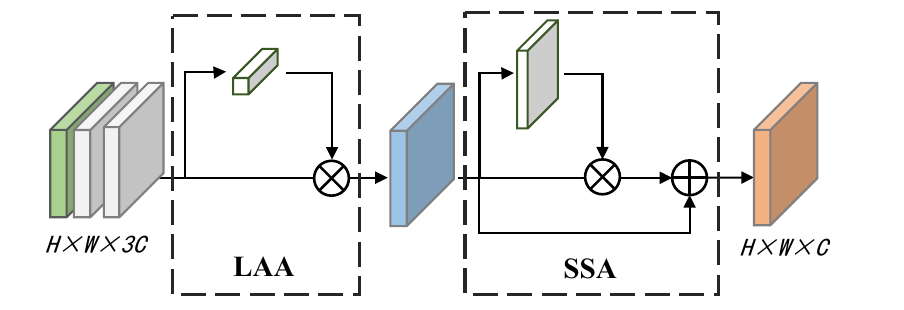}
  \end{center}
  \caption{Illustration of LDAM. LDAM contains a layer attention aggregation (LAA) and a simple spatial attention (SSA) to extract decoupled features for classification and localization.}3
  \label{fig:ldam}
\end{figure}

\par
Inspired by TOOD \cite{tood}, we present a lightweight decoupled attention module (LDAM) to decompose the localization and classification tasks with few extra parameters. LDAM includes a layer attention aggregation (LAA) and a simple spatial attention (SSA). The illustration of proposed LDAM is shown in Fig. \ref{fig:ldam}. In LAA, we first concatenate $X^{fpn}$ and $X^{conv}$ to get $X^{cat}\in \mathbb{R} ^{H\times W\times \left( N+1 \right) C}$ and execute global average pooling on the concatenated features. After that, a $1\times1$ conv layer is applied to learn a layer attention map $w^{layer}\in \mathbb{R} ^{\left( N+1 \right) C}$. There is no activation function used here to reduce computation. Another $1\times1$ conv layer is to reduce the dimension from $(N+1)\times C$ to $C$ with the layer attention map multiplied in the meantime. As a result, the aggregated features $X^{LAA}\in \mathbb{R} ^{H\times W\times C}$ can be obtained. Then, a simple spatial attention (SSA) is presented to enhance spatial decomposition. The attention map is generated by a $7\times7$ conv layer on global average pooling and global max pooling maps from $X^{LAA}$. A residual path with layer scale \cite{layerscale} is used to maintain stability and speed up the convergence.

\par
Taking the decoupled features extracted by LDAM, the classification branch performs class-agnostic foreground prediction with an output dimension of $H\times W\times 1$, while the regression branch predicts four offsets and an angle $(l, t, r, b, \theta)$ of the oriented bounding boxes. During the training, to maximize the usage of samples, QOPN discards random sampling and uses Focal Loss \cite{retinanet} to reduce the weight of well-classified samples. Rotated GIoU loss is employed for box regression, since IoU-based loss can not only alleviate the inconsistency problem of loss function and evaluation metric, but also avoid the boundary problem caused by angle periodicity.

\par
In addition, the input scale of feature map is limited in QOPN to reduce the computational cost. The vanilla RPN take $\left\{P_{2}, P_{3}, P_{4}, P_{5}, P_{6}\right\}$ from FPN as input. However, it is proved that $P_2$ costs a large portion of the computations but makes fewer contributions, while the high-level features are more efficient \cite{centernet2}, \cite{querydet}. Although low-level features may helpful to the detection of small object, in FGOD tasks, high-resolution images with few tiny objects are more common. In this case, QOPN discards $P_2$ and generates proposals on $\left\{P_{3}, P_{4}, P_{5}, P_{6}, P_{7}\right\}$ like a one-stage detector. The stride of each level is 8, 16, 32, 64 and 128, respectively. With a larger stride, there are significantly fewer anchor-points set on the feature map. In this way, QOPN can contain more parameters without increment of FLOPs.

\subsection{Bilinear Channel Fusion Network} \label{sec:bcfn}
We present a Bilinear Channel Fusion Network (BCFN) to enhance the feature representation of proposals. The motivation stems from the fact that high-level features contain more semantic information while low-level features are more likely to respond to local texture and patterns \cite{visualizing}. In two-stage FGOD methods, the second stage requires not only the semantic information for precise box regression but also high-resolution spatial details for accurate fine-grained recognition. Despite the pathway of FPN, the feature representation is still not rich enough. In addition, features for both stages extracted from FPN without feature decoupling will be harmful to the fine-grained recognition due to the confusion between sub-tasks. By cross-layer fusion, Our BCFN can address above two problems in an effective and efficient way.

\par
\begin{figure}[!tbp]
  \begin{center}
    \includegraphics[width=0.45\textwidth]{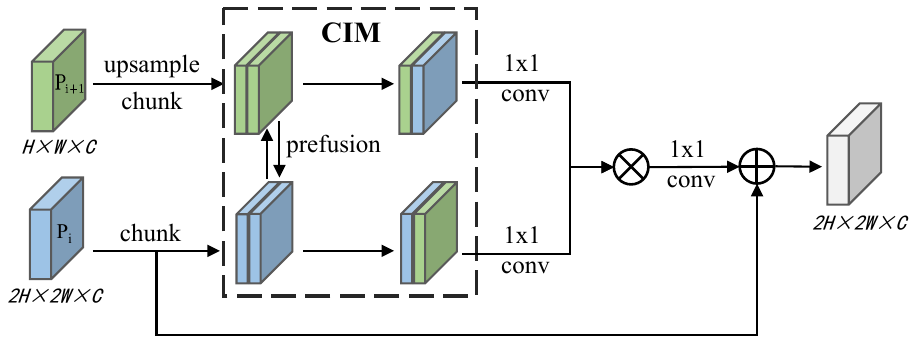}
  \end{center}
  \caption{Illustration of BCFN. BCFN contains a channel interaction module (CIM) for feature prefusion and the bilinear channel fusion implemented by 1 × 1 convolutions and a Hadamard product.}
  \label{fig:bcfn}
\end{figure}

As shown in Fig. \ref{fig:bcfn}, our BCFN takes two adjacent level feature maps as input to generate enhanced cross-level feature. In the beginning of BCFN, a channel interaction module (CIM) is proposed to fully harness the cross-level information via channel-wise prefusion. CIM performs channel replacement without new training parameters, which can be expressed as:
\begin{equation}
  \begin{aligned}
    \left[ X_{11},X_{12} \right] & =\mathrm{Chunk}\left( P_i \right)
    \\
    \left[ X_{21},X_{22} \right] & =\mathrm{Chunk}\left( \uparrow P_{i+1} \right)
    \\
    X_L                          & =\mathrm{Concat}\left( \left[ X_{11},X_{22} \right] \right)
    \\
    X_H                          & =\mathrm{Concat}\left( \left[ X_{21},X_{12} \right] \right)
    \\
  \end{aligned}
\end{equation}
where $P_i \in \mathbb{R} ^{2H\times 2W\times C}$ and $P_{i+1} \in \mathbb{R} ^{H\times W\times C}$ indicate the low-level and high-level feature maps. $H$, $W$ and $C$ represent the height, width, and the number of channels of the feature maps, respectively. $\mathrm{Chunk}\left( \cdot \right)$ is the channel chunk operation and $\mathrm{Concat}\left( \cdot \right)$ denotes the channel-wise concatenation operation. $\uparrow$ represents upsampling features via nearest neighbor interpolation.

\par
Based on the result of CIM, we perform the bilinear channel fusion (BCF) to accomplish the actual task of cross-level feature fusion. BCF draws inspiration from the bilinear pooling \cite{bilinear}, a classical method that has shown promising performance in the fine-grained image classification task. However, original bilinear pooling costs significant computational resources to calculate second-order features. To tackle this issue, we draw inspiration from the Multimodal Low-rank Bilinear Attention Networks (MLB) \cite{mlb} and implement the bilinear operation in BCF by two linear mappings and a Hadamard product, which can be formulated as:
\begin{equation}
  BCF\left( X_L,X_H,f,g,h \right) =h\left( f\left( X_L \right) \odot g\left( X_H \right) \right)
\end{equation}
where $ f\left( \cdot \right) $, $g\left( \cdot \right) $, and $h\left( \cdot \right) $ represent the channel-wise linear mappings implemented by 1 × 1 convolutions, respectively. The symbol $\odot$ denotes the Hadamard product between two matrices. BCF can also be deemed as a bilinear variant of the Gated Linear Unit (GLU) \cite{glu}, \cite{gluvariants}, which has been proved effective in natural language processing. The main difference between BCF and MLB/GLU lies in the fact that all operations in our design are channel-wise, since BCFN is applied prior to RoI Pooling, which can extract strong spatial features. In this case, there is no need to consume additional compute resources to prematurely perform spatial operations before focusing on the regions of interest.

\par
Considering that low-level features play a more important role in FGOD tasks, we add the original low-level feature map to the result as a shortcut. After performing the above operations, the bilinear fusion feature map $B_i \in \mathbb{R} ^{2H\times 2W\times C}$ can be obtained. Specifically, $\left\{P_{2}, P_{3}, P_{4}, P_{5}, P_{6}, P_{7}\right\}$ from FPN are introduced to BCFN and $\left\{B_{2}, B_{3}, B_{4}, B_{5}, B_{6} \right\}$ with the stride of $\left\{4, 8, 16, 32, 64 \right\}$ will be generated to support the next stage with feature enriched and decoupled.

\subsection{Adaptive Recognition Loss} \label{sec:arl}
To take full advantage of high-quality proposals, We propose the Adaptive Recognition Loss (ARL) to mainly address two problems existing in current two-stage FGOD methods. First, the vanilla R-CNN head ignore the quality variance of different proposals. We believe that high-quality proposals with higher foreground probability and more precise localization should be prioritized in the FGOD tasks. Second, the non-maximum suppression (NMS) of proposals has a negative impact on the fine-grained recognition. Due to the misalignment of classification and localization in the first stage, high-scoring proposals may not have accuracy bounding box. As a result, well-localized proposals might be removed. Besides, rotated NMS is time-consuming operation. The substitute horizontal NMS sometimes leads to wrongly removal, especially confronting the densely arranged proposals with a large aspect ratio.

\par
Our ARL is designed to address above issues based on Focal Loss \cite{retinanet}. The vanilla Focal Loss can be formulated as:
\begin{equation}
  \mathrm{FL}\left( p,y \right) =\begin{cases}
    -\alpha \left( 1-p \right) ^{\gamma}\log \left( p \right) , & \mathrm{if} \ y=1  \\
    -p^{\gamma}\log \left( 1-p \right) ,                        & \mathrm{otherwise} \\
  \end{cases}
\end{equation}
where $\alpha$ and $\gamma$ are two hyperparameters. $\alpha$ is used to balance the contribution of positive and negative samples while $\gamma$ adjusts the rate of focusing on hard examples. However, Focal Loss treats positive and negative samples equally, while positive samples are more valuable for FGOD. In addition, Focal Loss modulates by cross-entropy, ignoring the actual factors which affect fine-grained recognition. In that case, we design a new reweighing strategy for ARL. The joint measurement $t$ is proposed to re-weight positive samples by their quality, which can be calculated as:
\begin{equation}
  t=\sqrt{s\times q}
\end{equation}
where $s$ denotes the classification score of the proposal provided by QOPN. Compared with the vanilla RPN which places most emphasis on the recall, our QOPN can generate more reliable scores to precisely measure the quality of the proposal in terms of the foreground probability. $q$ is the output IoU after box regression in the second stage, which indicates the quality according to localization. Overall, ARL is able to jointly assess the quality of each proposal based on the classification from the first stage and localization from the second stage. Then our ARL can be formulated as:
\begin{equation}
  \mathrm{ARL}\left( p,y \right) =\begin{cases}
    -te^{\beta t}\log \left( p \right) , & \mathrm{if} \ y=1  \\
    -p^{\gamma}\log \left( 1-p \right) , & \mathrm{otherwise} \\
  \end{cases}
\end{equation}
where the hyperparameter $\beta$ controls the variance of different proposals. Comparing with Focal Loss, the hyperparameter $\alpha$ is removed in our ARL, since $\beta$ can also contribute to adjust the overall weight of positive samples. With ARL, we no longer perform NMS on the proposals but retain noisy proposals to avoid the absence of high-quality proposals. Discarding NMS also contributes much to inference speed. Besides, since ARL can dynamically increase the weight of high-quality proposals, we do not sample proposals during the training to maximize the utilization of all high-quality proposals.

\section{Experimental Results and Analysis} \label{sec:result}
In this section, we design extensive experiments to evaluate the performance of our PETDet on multiple FGOD datasets. We compare our methods with the state-of-the-art oriented object detection methods to demonstrate our superiority. Besides, We conduct extensive experiments to verify the effectiveness of each module and the optimal parameter settings.

\subsection{Datasets}
Multiple datasets are adopted to evaluate the effectiveness and robustness of our method comprehensively, including the FAIR1M \cite{fair1m} dataset, the MAR20 \cite{mar20} dataset and the ShipRSImageNet \cite{shiprs} dataset. Our mainly experimental results are performed on FAIR1M, the largest multi-class FGOD dataset. The MAR20 dataset and ShipRSImageNet dataset are single-class FGOD dataset, producing for aircraft and ship detection and recognition respectively. We adopt these three datasets to comprehensively evaluate our method in both multi-class and single-class FGOD tasks.

\subsubsection{FAIR1M Dataset}
FAIR1M \cite{fair1m} is currently the largest remote sensing fine-grained object detection dataset. The image size ranges from 1000 × 1000 to 10,000 × 10,000 pixels. All objects in the dataset are annotated with respect to 5 categories (Airplane, Ship, Vehicle, Court,and Road) and 37 sub-categories with OBB annotations. Except 3 other categories (other-airplane, other-ship, and other-vehicle), there are 34 fine-grained classes in FAIR1M, i.e., Boeing 737 (B737), Boeing 777 (B777), Boeing 747 (B747), Boeing 787 (B787), Airbus A320 (A320), Airbus A220 (A220), Airbus A330 (A330), Airbus A350 (A350), COMAC C919 (C919), COMAC ARJ21 (ARJ21), passenger ship (PS) motorboat (MB), fishing boat (FB), tugboat (TB), engineering ship (ES), liquid cargo ship (LCS), dry cargo ship (DCS), warship (WS), small car (SC), bus (BUS), cargo truck (CT), dump truck (DT), van (VAN), trailer (TRI), tractor (TRC), truck tractor (TT), excavator (EX), baseball field (BF), basketball court (BC), football field (FF), tennis court (TC), roundabout (RA), intersection (IS), and bridge (BR).
\par
The FAIR1M dataset consists of two versions: FAIR1M-v1.0 and FAIR1M-v2.0. The 1.0 version contains 16488 images for training and 8137 images for testing, respectively. Comparing with FAIR1M-v1.0, the 2.0 version introduces an extra validation set and an expanded testing set, while the train set of them are consistent.

\subsubsection{MAR20 Dataset}
MAR20 \cite{mar20} is a dataset of remote sensing military aircraft recognition. It consists of 3842 images, including 1311 images for training and 2511 images for test. The size of images is mostly 800 × 800 pixels. In MAR20, there are 22341 aircraft instances with 20 different types gathered from 60 military airports across the United States, Russia, and other countries using Google Earth. All instance has both horizontal bounding box and oriented bounding box annotations.

\subsubsection{ShipRSImageNet Dataset}
ShipRSImageNet \cite{shiprs} is a large-scale fine-grained ship detection datasets. Most of the data are collected from Google Earth and supplemented by HRSC2016. The resolution of ShipRSImageNet ranges from 0.12 to 6 m, and the size of images ranges from 930 × 930 to 1024 × 1024 pixels. The dataset contains over 3435 images with 17,573 ship instances. Instances are annotated with HBB, OBB and polygons annotations. In ShipRSImageNet, ships are hierarchically classified into four levels. We mainly evaluate models on level 3, where ship objects are divided into 50 fine-grained classes at type level.

\subsection{Implementation Details}
We implement our PETDet and other comparison models by the mmrotate \cite{mmrotate} toolbox. All models are trained on 4 NVIDIA GeForce RTX3090 GPUs with total batch size set to 8 (2 images per GPU). We use SGD with momentum of 0.9 and the weight decay of 0.0001 as default optimization parameters. The initial learning rate in all experiments is 0.02. No data augmentation except random flipping is used during the training.

\par
It should be noted that some experimental settings vary across different datasets. For FAIR1M, we use training and validation sets for training and the rest for testing. All images are cropped into 1024 × 1024 patches with an overlap of 200. And models are trained models with 12 epochs and the learning rate is reduced at the factor of 0.1 in the 8th and 11th. For MAR20 and ShipRSImageNet, the official train-test split is adopted. Models are trained with longer 36 epochs and the learning rate is reduced at the factor of 0.1 in the 24th and 33rd. The input image size for MAR20 and ShipRSImageNet is set as 800 × 800 and 1024 × 1024, respectively.

\subsection{Evaluation Metric}
We choose mean average precision (AP) as the major evaluation metric. AP of each class is calculated based on precision (P) and recall (R), which are calculated by:
\begin{equation}
  Precision=\frac{TP}{TP+FP}
\end{equation}
\begin{equation}
  Recall=\frac{TP}{TP+FN}
\end{equation}
where TP, FP, and FN respectively represent the true-positive, false-positive and false-negative. The definitions of positive and negative samples depended on the IoU threshold, e.g, $\mathrm{AP_{50}}$ employs the IoU = 0.5 as the threshold. By set different confidence thresholds to change the recall, the P–R curve composed of a series of different P and R can be obtained. After that, AP can be calculated by the P–R curve of each class.

\par
Note that some details of AP calculation are not fully same for different datasets. For FAIR1M, Only $\mathrm{AP_{50}}$ will be reported by the online evaluation server according to Pascal VOC 2012 metric. Thirty-four fine-grained categories are used for calculation, and the accuracy of three other categories (other-airplane, other-ship, and other-vehicle) is not included. Conversely, for MAR20 and ShipRSImageNet, $\mathrm{AP_{50}}$ is calculated by Pascal VOC \cite{voc} 2007 metric. Besides, $\mathrm{AP_{75}}$ and $\mathrm{AP_{50:95}}$ are calculated to dissect the quality of localization. Unless otherwise specified, all \textit{AP} in our paper by default stand for $\mathrm{AP_{50}}$ rather than $\mathrm{AP_{50:95}}$ for greater focus on fine-grained recognition performance.

\par
In assessing the efficacy of our QOPN, the average recall (AR) with different IoU threshold and proposals number is used to evaluate the quality of generated proposals. We also evaluate the speed of our PETDet and other mainstream oriented object detectors. The FPS result is reported from experiments on single RTX 3090 GPU. The time of post-processing (e.g., NMS) is included.

\begin{table}[!t]
  \centering
  \caption{Quantitative results on FAIR1M-2.0 dataset. * indicates the result is from \cite{sfr}. The others are from our re-implementation.}
  \begin{tblr}{
      stretch=1.1,
      column{2-5} = {c},
      cell{2}{1} = {c=5}{l},
      cell{9}{1} = {c=5}{l},
      hline{3} = {-}{},
      hline{9} = {1}{-}{},
      hline{10} = {-}{},
      vline{3-4} = {-}{},
    }
    \toprule
    Method                          & Reference & Backbone   & $\mathrm{AP_{50}}$ & FPS           \\
    \midrule
    \textit{One-stage methods:}                                                                   \\
    RetinaNet \cite{retinanet}      & ICCV2017  & R-50-FPN   & 26.07              & 24.7          \\
    GWD \cite{orcnn}                & ICML2021  & R-50-FPN   & 31.48              & —             \\
    KLD \cite{kld}                  & NIPS2021  & R-50-FPN   & 32.12              & —             \\
    R$^3$Det \cite{r3det}           & AAAI2021  & R-50-FPN   & 36.04              & 18.5          \\
    S$^2$ANet \cite{s2anet}         & TGRS2021  & R-50-FPN   & 30.93              & 21.5          \\
    FCOS \cite{fcos}                & CVPR2019  & R-50-FPN   & 39.98              & \textbf{28.1} \\
    \textit{Two-stage methods:}                                                                   \\
    Faster R-CNN \cite{faster-rcnn} & CVPR2018  & R-50-FPN   & 41.64              & 24.7          \\
    Gliding Vertex \cite{GV}        & TPAMI2020 & R-50-FPN   & 41.27              & 24.1          \\
    RoI Transformer \cite{roitrans} & CVPR2019  & R-50-FPN   & 44.03              & 21.7          \\
    Orinted R-CNN \cite{orcnn}      & ICCV2021  & R-50-FPN   & 43.90              & 23.4          \\
    ReDet \cite{redet}              & CVPR2021  & ReR-50-FPN & 46.03              & 15.8          \\
    SFRNet* \cite{sfr}              & TGRS2023  & R-50-FPN   & 45.68              & —             \\
    \textbf{PETDet (Ours)}          & —         & R-50-FPN   & \textbf{48.81}     & 23.3          \\
    \bottomrule
  \end{tblr}
  \label{tab:fair2.0}
\end{table}

\begin{table*}[!t]
  \tiny
  \centering
  \caption{Comparison of per-class performance on FAIR1M-2.0. Red numbers are the optimal results and blue numbers represent the suboptimal results.}
  \SetTblrInner{rowsep=1pt,colsep=0.5pt}
  \begin{tblr}{
    width=1.0\textwidth,
    colspec={X[0.8,c]X[0.7,c]X[1.05,c]X[c]X[c]X[c]X[c]X[c]X[c]X[1.15,c]X[c]X[c]X[c]X[1.2,c]},
    cells = {c},
    cell{1}{1} = {c=2}{},
    cell{2}{1} = {c=2}{},
    cell{3}{1} = {r=10}{},
    cell{13}{1} = {r=8}{},
    cell{21}{1} = {r=9}{},
    cell{30}{1} = {r=4}{},
    cell{34}{1} = {r=3}{},
    cell{37}{1} = {c=2}{},
    vline{2} = {1-2,37}{},
    vline{2-3} = {1-37}{},
    vline{3} = {4-12,14-20,22-29,31-33,35-36}{},
    hline{1,2-3,13,21,30,34,37,38} = {-}{},
      }
    Method             &       & RetinaNet-O \cite{retinanet} & GWD \cite{gwd} & KLD\cite{kld} & R3Det \cite{r3det} & S2ANet \cite{s2anet} & FCOS-O \cite{fcos} & FRCNN-O \cite{faster-rcnn} & Gliding Vertex \cite{GV} & RoI Trans \cite{roitrans} & ReDet \cite{redet} & ORCNN \cite{orcnn} & \textbf{PETDet (Ours)} \\
    Backbone           &       & R-50                         & R-50           & R-50          & R-50               & R-50                 & R-50               & R-50                       & R-50                     & R-50                      & ReR-50             & R-50               & R-50                   \\
    Airplane           & B737  & 37.60                        & 38.05          & 38.82         & 39.28              & 39.74                & 38.12              & 42.29                      & 41.36                    & 46.52                     & 44.02              & \bluebf{44.16}     & \redbf{49.96}          \\
                       & B747  & 86.54                        & 87.59          & 88.53         & 87.47              & 88.74                & 89.30              & 92.73                      & 93.02                    & 93.05                     & \redbf{94.41}      & 93.13              & \bluebf{93.60}         \\
                       & B777  & 28.21                        & 28.66          & 31.68         & 31.33              & 33.51                & 31.65              & 35.68                      & 35.67                    & 41.60                     & \bluebf{42.89}     & 39.64              & \redbf{43.77}          \\
                       & B787  & 48.91                        & 50.52          & 55.97         & 53.93              & 52.09                & 54.72              & 60.36                      & 58.94                    & 62.88                     & 60.17              & \bluebf{61.41}     & \redbf{67.27}          \\
                       & C919  & 0.44                         & 0.35           & 0.35          & 0.70               & 0.50                 & 0.62               & 8.90                       & 2.13                     & \bluebf{9.90}             & 9.29               & 2.25               & \redbf{34.53}          \\
                       & A220  & 51.19                        & 52.37          & 50.12         & 51.77              & 49.34                & 52.88              & 54.03                      & 53.51                    & 53.80                     & \bluebf{56.93}     & 54.53              & \redbf{59.91}          \\
                       & A321  & 65.89                        & 68.87          & 69.13         & 69.87              & 70.52                & 69.63              & 68.34                      & 67.03                    & 71.07                     & \bluebf{71.81}     & 70.61              & \redbf{76.72}          \\
                       & A330  & 27.86                        & 33.79          & 40.78         & 44.99              & 47.54                & 50.31              & 58.69                      & 58.21                    & 61.48                     & \bluebf{62.98}     & 59.94              & \redbf{64.33}          \\
                       & A350  & 50.82                        & 48.41          & 49.39         & 55.16              & 47.07                & 58.98              & 67.67                      & 66.14                    & 67.02                     & \redbf{73.93}      & 67.54              & \bluebf{73.29}         \\
                       & ARJ21 & 3.47                         & 3.34           & 3.82          & 2.75               & 2.20                 & 3.15               & 9.97                       & 10.10                    & 14.35                     & \redbf{17.42}      & 11.97              & \bluebf{15.29}         \\
    Ship               & PS    & 1.92                         & 4.76           & 4.56          & 9.62               & 4.80                 & 11.28              & 12.00                      & 13.06                    & 12.71                     & \redbf{15.52}      & \bluebf{15.20}     & 13.43                  \\
                       & MB    & 10.74                        & 36.16          & 35.69         & 38.26              & 23.47                & 54.12              & 53.17                      & 54.81                    & 58.74                     & 61.36              & 59.62              & \redbf{64.15}          \\
                       & FB    & 1.43                         & 5.51           & 5.80          & 9.55               & 4.67                 & 16.19              & 16.99                      & 15.55                    & 19.33                     & \redbf{29.59}      & \bluebf{26.46}     & 24.94                  \\
                       & TB    & 12.87                        & 21.55          & 22.22         & 25.39              & 15.62                & 29.83              & 29.38                      & 29.34                    & \redbf{30.79}             & 30.71              & \bluebf{30.23}     & 30.05                  \\
                       & ES    & 4.32                         & 12.29          & 11.91         & 10.44              & 7.59                 & 13.97              & 13.00                      & 12.78                    & 13.81                     & \redbf{20.22}      & 15.92              & \bluebf{17.82}         \\
                       & LCS   & 8.33                         & 17.98          & 17.35         & 46.48              & 31.81                & 46.74              & 43.29                      & 43.76                    & 46.96                     & \bluebf{49.78}     & 49.06              & \redbf{50.67}          \\
                       & DCS   & 14.11                        & 28.86          & 28.96         & 45.84              & 33.34                & 49.55              & 44.45                      & 44.08                    & 50.52                     & \redbf{52.59}      & 51.12              & \bluebf{51.18}         \\
                       & WS    & 1.21                         & 5.36           & 5.77          & 22.32              & 7.16                 & 28.82              & 20.27                      & 20.23                    & 27.33                     & \redbf{33.18}      & \bluebf{32.37}     & 32.04                  \\
    Vehicle            & SC    & 30.40                        & 46.80          & 49.61         & 51.17              & 47.77                & 54.92              & 53.78                      & 53.32                    & 58.10                     & \bluebf{60.00}     & 56.87              & \redbf{72.00}          \\
                       & BUS   & 2.69                         & 5.79           & 7.08          & 3.95               & 2.38                 & 12.40              & 25.68                      & 31.18                    & 31.76                     & 23.88              & \bluebf{30.46}     & \redbf{43.51}          \\
                       & CT    & 8.49                         & 25.45          & 23.62         & 34.27              & 19.75                & 44.78              & 47.36                      & 47.42                    & 49.76                     & 49.07              & \bluebf{49.95}     & \redbf{55.60}          \\
                       & DT    & 10.09                        & 19.59          & 17.67         & 27.98              & 24.10                & 34.71              & 45.69                      & 45.94                    & 48.78                     & \bluebf{49.52}     & 48.82              & \redbf{54.76}          \\
                       & VAN   & 20.81                        & 38.69          & 41.49         & 43.50              & 39.53                & 51.71              & 47.87                      & 47.01                    & 52.92                     & \bluebf{55.50}     & 51.39              & \redbf{69.83}          \\
                       & TRI   & 0.00                         & 0.12           & 0.08          & 10.43              & 0.08                 & 12.19              & 12.17                      & 13.12                    & 13.68                     & 13.45              & \redbf{15.99}      & \bluebf{14.24}         \\
                       & TRC   & 0.00                         & 0.00           & 0.04          & 0.26               & 0.08                 & 1.24               & 1.58                       & 1.02                     & 1.81                      & 1.33               & \bluebf{1.85}      & \redbf{1.94}           \\
                       & EX    & 0.16                         & 0.11           & 0.42          & 4.84               & 1.71                 & 13.39              & 11.15                      & 13.22                    & 13.21                     & 14.28              & \redbf{15.26}      & \bluebf{14.82}         \\
                       & TT    & 0.00                         & 0.08           & 0.03          & 0.42               & 0.02                 & 22.79              & 13.45                      & 11.05                    & 16.34                     & \bluebf{21.20}     & 5.39               & \redbf{33.14}          \\
    Court              & BC    & 30.54                        & 38.45          & 38.23         & 47.66              & 40.69                & 50.64              & 55.38                      & 54.34                    & 56.90                     & \redbf{64.65}      & \bluebf{57.56}     & 54.88                  \\
                       & TC    & 74.59                        & 80.85          & 83.93         & 85.68              & 79.13                & 86.89              & 88.40                      & 87.51                    & \bluebf{88.88}            & \redbf{89.35}      & 88.56              & 88.17                  \\
                       & FF    & 54.73                        & 59.55          & 61.54         & 59.33              & 46.33                & 58.43              & 60.23                      & 60.22                    & 62.80                     & \bluebf{66.89}     & 63.34              & \redbf{67.56}          \\
                       & BF    & 87.28                        & 89.68          & 90.31         & 90.25              & 87.14                & 89.04              & 91.47                      & 90.01                    & 90.81                     & \bluebf{91.74}     & 90.90              & \redbf{92.36}          \\
    Road               & IS    & 60.07                        & 62.27          & 60.32         & 61.23              & 56.07                & 65.88              & 65.04                      & 63.41                    & 64.40                     & \bluebf{65.20}     & 64.57              & \redbf{65.75}          \\
                       & RA    & 35.13                        & 32.63          & 33.21         & 33.17              & 29.15                & 27.51              & 33.09                      & 33.20                    & 31.90                     & \bluebf{32.84}     & 31.24              & \redbf{35.73}          \\
                       & BR    & 15.46                        & 25.91          & 23.72         & 25.97              & 18.02                & 32.89              & 32.18                      & 31.38                    & 32.99                     & \redbf{39.25}      & \bluebf{35.36}     & 32.41                  \\
    $\mathrm{AP_{50}}$ &       & 26.07                        & 31.48          & 32.12         & 36.04              & 30.93                & 39.98              & 41.64                      & 41.27                    & 44.03                     & \bluebf{46.03}     & 43.90              & \redbf{48.81}
  \end{tblr}
  \label{tab:fair2.0fg}
\end{table*}

\begin{figure*}[!tbp]
  \centering
  \begin{minipage}[c]{.02\textwidth}
    \rotatebox{90}{(a) gt}
  \end{minipage}
  \begin{minipage}[c]{.157\textwidth}
    \centering
    \includegraphics[width=\textwidth]{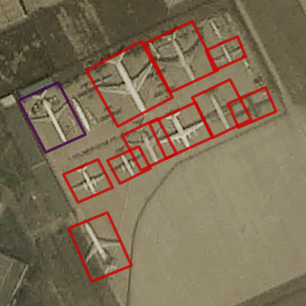}
  \end{minipage}
  \begin{minipage}[c]{.157\textwidth}
    \centering
    \includegraphics[width=\textwidth]{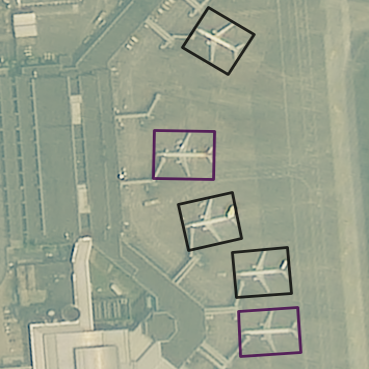}
  \end{minipage}
  \begin{minipage}[c]{.157\textwidth}
    \centering
    \includegraphics[width=\textwidth]{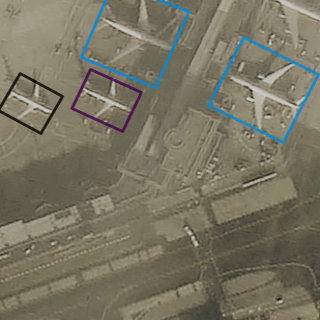}
  \end{minipage}
  \begin{minipage}[c]{.157\textwidth}
    \centering
    \includegraphics[width=\textwidth]{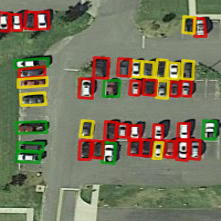}
  \end{minipage}
  \begin{minipage}[c]{.157\textwidth}
    \centering
    \includegraphics[width=\textwidth]{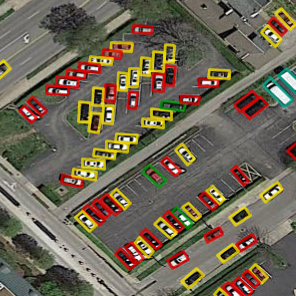}
  \end{minipage}
  \begin{minipage}[c]{.157\textwidth}
    \centering
    \includegraphics[width=\textwidth]{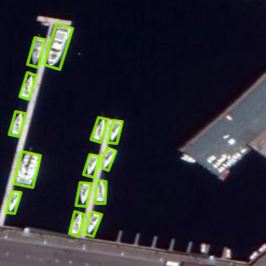}
  \end{minipage}

  \vspace{3pt}
  \centering
  \begin{minipage}[c]{.02\textwidth}
    \rotatebox{90}{(b) baseline}
  \end{minipage}
  \begin{minipage}[c]{.157\textwidth}
    \centering
    \includegraphics[width=\textwidth]{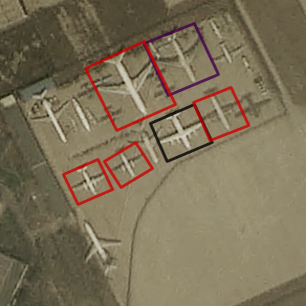}
  \end{minipage}
  \begin{minipage}[c]{.157\textwidth}
    \centering
    \includegraphics[width=\textwidth]{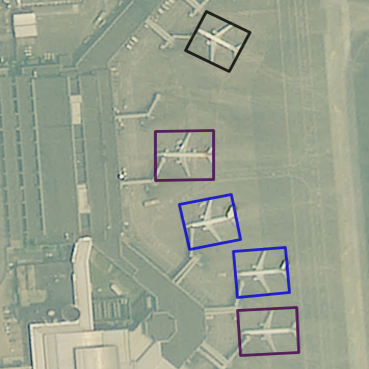}
  \end{minipage}
  \begin{minipage}[c]{.157\textwidth}
    \centering
    \includegraphics[width=\textwidth]{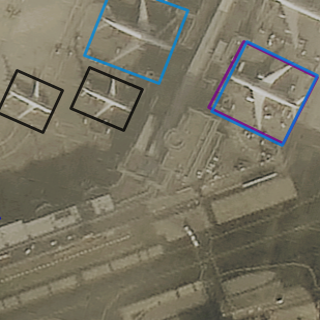}
  \end{minipage}
  \begin{minipage}[c]{.157\textwidth}
    \centering
    \includegraphics[width=\textwidth]{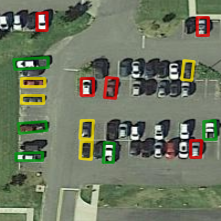}
  \end{minipage}
  \begin{minipage}[c]{.157\textwidth}
    \centering
    \includegraphics[width=\textwidth]{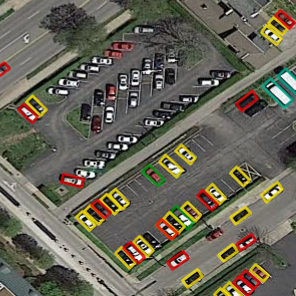}
  \end{minipage}
  \begin{minipage}[c]{.157\textwidth}
    \centering
    \includegraphics[width=\textwidth]{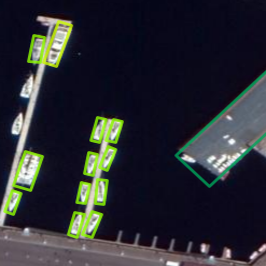}
  \end{minipage}

  \vspace{3pt}
  \centering
  \begin{minipage}[c]{.02\textwidth}
    \rotatebox{90}{(c) PETDet}
  \end{minipage}
  \begin{minipage}[c]{.157\textwidth}
    \centering
    \includegraphics[width=\textwidth]{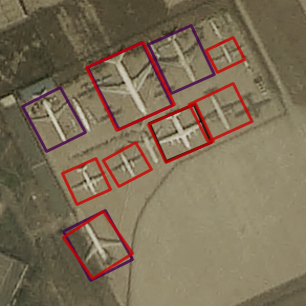}
  \end{minipage}
  \begin{minipage}[c]{.157\textwidth}
    \centering
    \includegraphics[width=\textwidth]{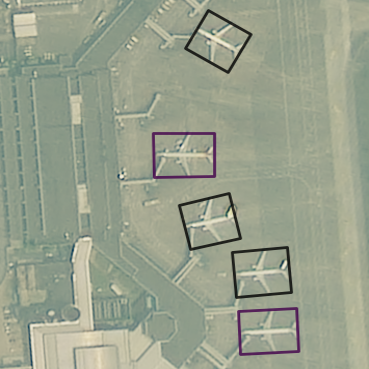}
  \end{minipage}
  \begin{minipage}[c]{.157\textwidth}
    \centering
    \includegraphics[width=\textwidth]{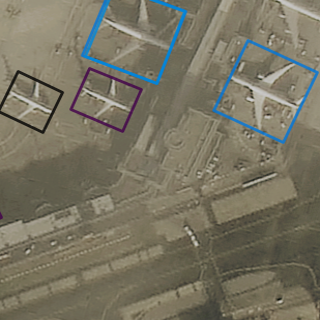}
  \end{minipage}
  \begin{minipage}[c]{.157\textwidth}
    \centering
    \includegraphics[width=\textwidth]{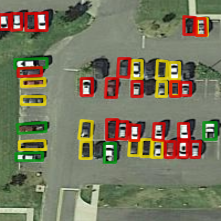}
  \end{minipage}
  \begin{minipage}[c]{.157\textwidth}
    \centering
    \includegraphics[width=\textwidth]{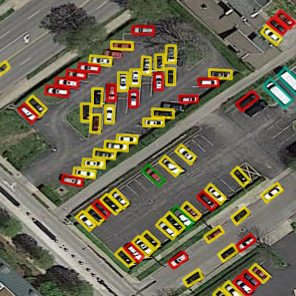}
  \end{minipage}
  \begin{minipage}[c]{.157\textwidth}
    \centering
    \includegraphics[width=\textwidth]{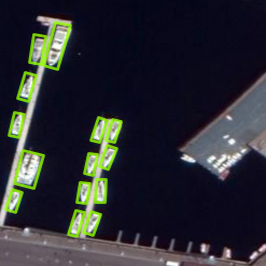}
  \end{minipage}

  \caption{Qualitative comparisons on FAIR1M-v2.0 validation. Ground-truth annotations shown in the top row as references. The results in the second row and the last row respectively from Oriented R-CNN (baseline) and our proposed PETDet. Different colored bounding boxes represent objects of different fine-grained categories. We set the threshold as 0.5 to filter out bounding boxes with low confidence here.}
  \label{fig:qualitative_result}
\end{figure*}

\subsection{Comparison with State-of-the-Art Methods}
In this section, we compare PETDet with more than ten mainstream rotated object detection methods, including single-stage and two-stage methods. Experimental results show that our PETDet achieves state-of-the-art performance and outperform previous method lots on multiple datasets.

\subsubsection{Quantitative and Qualitative Result on FAIR1M}
For FAIR1M, the primary experimental results are performed on FAIR1M-v2.0. We report 11 mainstream oriented object detectors for comparison, including advanced two-stages methods such as Oriented R-CNN, ReDet, etc. The quantitative results are shown in Table \ref{tab:fair2.0}, with the best performance highlighted in bold. All methods except ReDet use ResNet50 \cite{resnet} as the backbone by default. We did not report the FPS for GWD and KLD as they merely replace the regression loss based on RetinaNet. As can be seen, our PETDet largely surpasses all current oriented detectors by a clear margin with 48.81 $\mathrm{AP_{50}}$. Under same experiment settings, PETDet outperforms baseline Oriented R-CNN by 4.91 AP and the previous best RoI Transformer by 4.78 AP, respectively. PETDet also can outperform ReDet with the vanilla ResNet50, while ReDet uses heavier ReResNet50 for rotation equivariant. Compared with the baseline Oriented R-CNN, our PETDet maintains a comparable inference speed. Additionally, the $\mathrm{AP_{50}}$ for each specific fine-grained category is presented in Table \ref{tab:fair2.0fg}. It shows that for some challenging categories such as C919 and truck tractor (TT), PETDet has a notable advantage comparing with other methods.

\begin{figure*}[!tbp]
  \centering
  \begin{minipage}[b]{.36\textwidth}
    \centering
    \includegraphics[width=\textwidth]{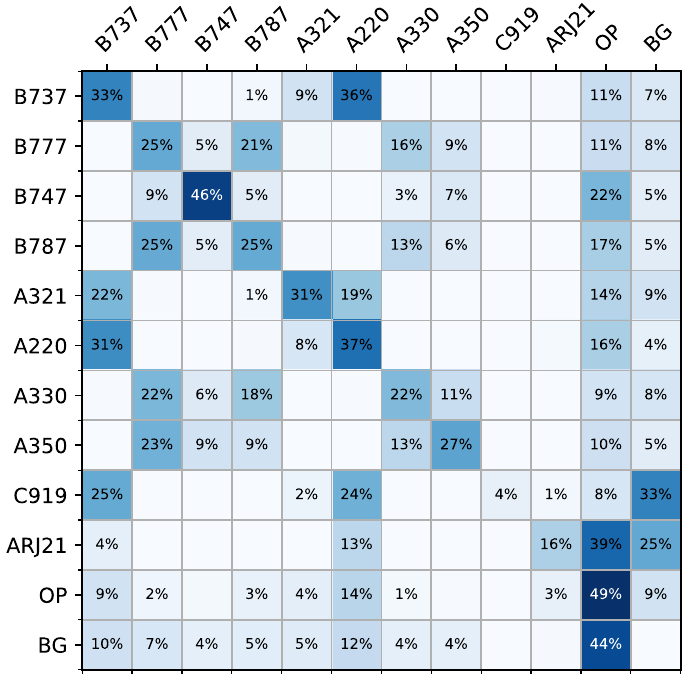}
  \end{minipage}%
  \begin{minipage}[b]{.28\textwidth}
    \centering
    \includegraphics[width=\textwidth]{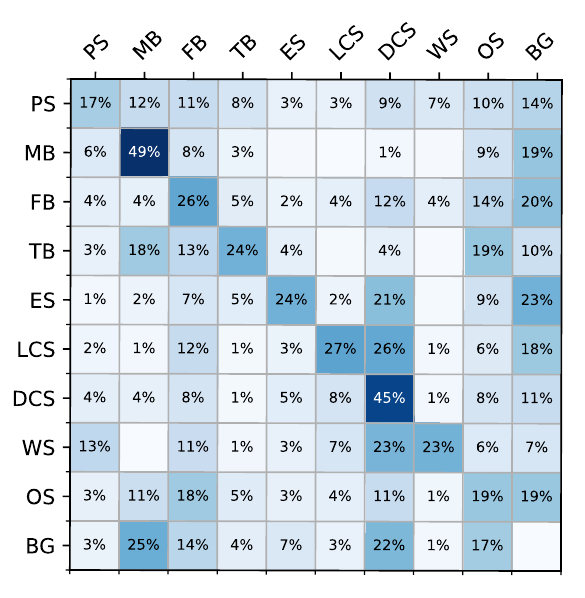}
  \end{minipage}
  \begin{minipage}[b]{.32\textwidth}
    \centering
    \includegraphics[width=\textwidth]{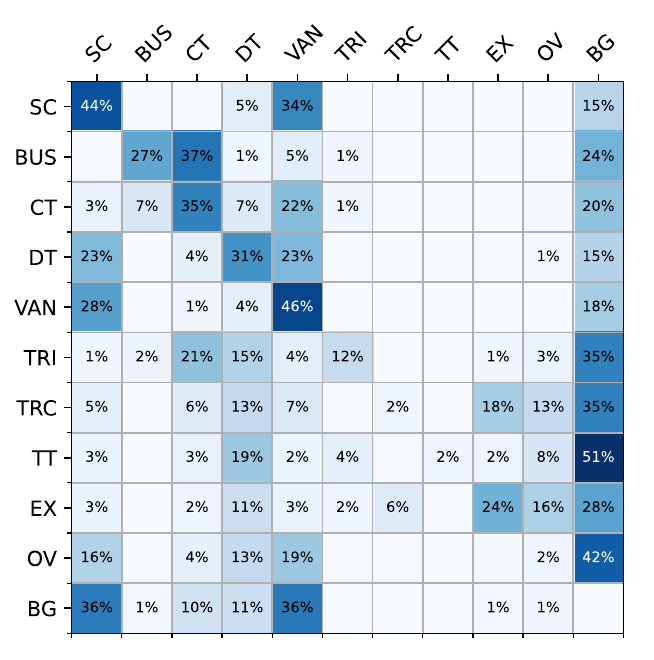}
  \end{minipage}

  \begin{minipage}[b]{.36\textwidth}
    \centering
    \includegraphics[width=\textwidth]{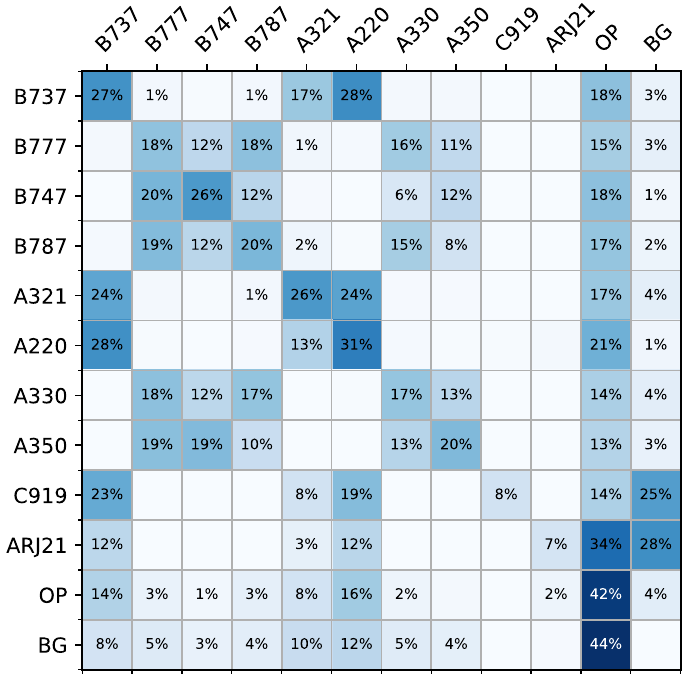}
    \vspace{-1.5\baselineskip}
    \subcaption{}\label{fig_cm_airplane}
  \end{minipage}%
  \begin{minipage}[b]{.28\textwidth}
    \centering
    \includegraphics[width=\textwidth]{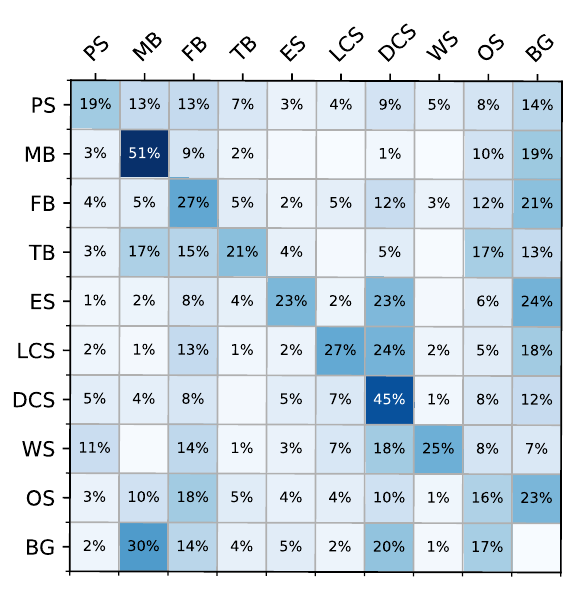}
    \vspace{-1.5\baselineskip}
    \subcaption{}\label{fig_cm_ship}
  \end{minipage}
  \begin{minipage}[b]{.32\textwidth}
    \centering
    \includegraphics[width=\textwidth]{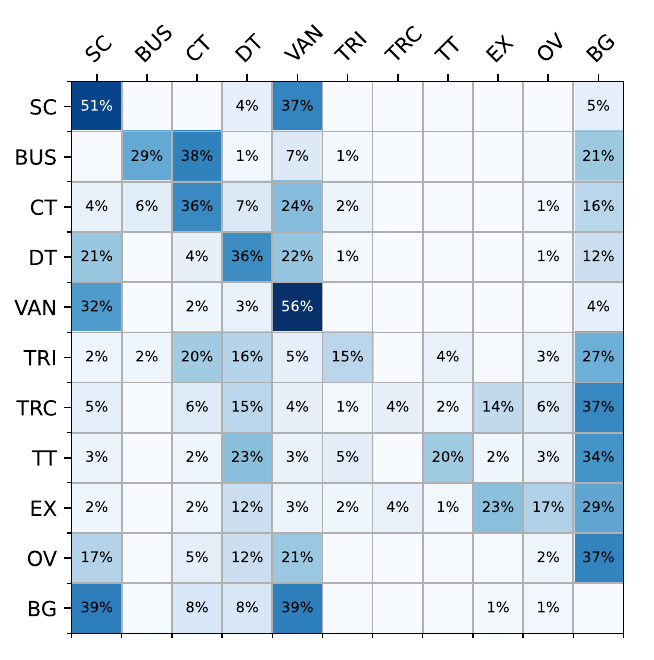}
    \vspace{-1.5\baselineskip}
    \subcaption{}\label{fig_cm_vehicle}
  \end{minipage}

  \caption{Confusion matrices of detection results obtained from Oriented R-CNN (top) and our PETDet (bottom). The horizontal and vertical coordinates of the matrices represent ground truths and the predictions, respectively. BG refers to the background. The vacancy values are nearly zero. (a) Airplane. (b) Ship. (c) Vehicle.}
  \label{fig:cm}%
\end{figure*}

Fig. \ref{fig:cm} illustrates the confusion matrices on FAIR1M-v2.0 validation, which are calculated based on the detection results obtained from the baseline Oriented R-CNN and our PETDet. It suggests that our PETDet outperforms the baseline in most fine-grained classes of ships and vehicles but appears to perform worse in airplane recognition. However, PETDet can achieve a higher AP than Oriented R-CNN in almost all categories. The variance is primarily attributable to the differences between metric calculations. AP is a ranked-base metric which takes both precision and recall into account, while the confusion matrix only focuses on precision with the confidence score ignored. In that case, PETDet tends to make more false-positive predictions with low confidence scores, resulting in a reduced proportion of true positives. However, these low-confidence predictions can be removed by setting a certain threshold in practice.

\par
Then, we have also analyzed the qualitative result on FAIR1M-v2.0 validation. Fig. \ref{fig:qualitative_result} gives the visualizations of fine-grained detection results obtained by Oriented R-CNN and our PETDet. It can be revealed that our proposal enhancement strategy can reduce false negatives in PETDet, particularly for small objects. Meanwhile, the emphasis on high-quality samples also facilitates fine-grained recognition. Without contrastive learning, our PETDet can also predict fine-grained categories closer to ground truth than the baseline Oriented R-CNN.

\par
Table \ref{tab:fair1.0} reports the quantitative results on FAIR1M-1.0. Note that the results reported by PCLDet are based on a score threshold of 0.001, instead of 0.05 in our default settings. So we reimplement PCLDet for fair comparison. Other results are cited from \cite{sfr}. Though all methods use ResNet50 as the backbone, some methods such as DAL, RIDet, CFC-Net, and TIOE-Det adopted more relaxing settings for data augmentation or training schedule. And some comparison methods are trained with a smaller batchsize, which may bring extra improvement to the results. The experimental results show that PETDet surpasses previous one-stage and two-stage oriented object detectors on FAIR1M-v1.0. Comparing with contrast learning based methods PCLDet and SFRNet, our proposed PETDet outperforms them by 2.71 AP and 2.22 AP respectively. Both experiments on FAIR1M-v2.0 and FAIR1M-v1.0 show that our PETDet can achieve significant improvements in multi-class FGOD tasks.

\begin{table}[!t]
  \centering
  \caption{Comparison results on FAIR1M-1.0. * indicates our re-implementation. Other results for comparison are from \cite{sfr}.}
  \begin{tblr}{
      stretch=1.05,
      column{2-4} = {c},
      cell{2}{1} = {c=4}{l},
      cell{13}{1} = {c=4}{l},
      hline{3} = {-}{},
      hline{13} = {1}{-}{},
      hline{14} = {-}{},
      vline{3-4} = {-}{},
    }
    \toprule
    Method                          & Reference & Backbone & $\mathrm{AP_{50}}$ \\
    \midrule
    \textit{One-stage methods:}                                                 \\
    RetinaNet \cite{retinanet}      & ICCV2017  & R-50-FPN & 26.58              \\
    GWD \cite{gwd}                  & ICML2021  & R-50-FPN & 28.13              \\
    KLD \cite{kld}                  & NIPS2021  & R-50-FPN & 28.25              \\
    DAL \cite{dal}                  & AAAI2021  & R-50-FPN & 29.00              \\
    RIDet \cite{ridet}              & GRSL2021  & R-50-FPN & 31.58              \\
    R$^3$Det \cite{r3det}           & AAAI2021  & R-50-FPN & 31.07              \\
    S$^2$ANet \cite{s2anet}         & TGRS2021  & R-50-FPN & 34.71              \\
    FCOS \cite{fcos}                & CVPR2019  & R-50-FPN & 34.10              \\
    CFC-Net \cite{cfc}              & TGRS2022  & R-50-FPN & 34.31              \\
    TIOE-Det \cite{tioe}            & JPRS2023  & R-50-FPN & 35.16              \\
    \textit{Two-stage methods:}                                                 \\
    Fatser R-CNN \cite{faster-rcnn} & CVPR2018  & R-50-FPN & 36.83              \\
    Gliding Vertex \cite{GV}        & TPAMI2020 & R-50-FPN & 36.47              \\
    RoI Transformer \cite{roitrans} & CVPR2019  & R-50-FPN & 39.15              \\
    Orinted R-CNN \cite{orcnn}      & ICCV2021  & R-50-FPN & 38.85              \\
    PCLDets*                        & TGRS2023  & R-50-FPN & 40.25              \\
    SFRNet \cite{sfr}               & TGRS2023  & R-50-FPN & 40.74              \\
    \textbf{PETDet (Ours)}          & —         & R-50-FPN & \textbf{42.96}     \\
    \bottomrule
  \end{tblr}
  \label{tab:fair1.0}
\end{table}

\subsubsection{Quantitative Result on MAR20}
To further validate the effectiveness of PETDet on the single-class FGOD task, we conducted comparison experiments on the MAR20 dataset. MAR20 contains only aircraft targets, where detection difficulty is relatively lower compared to FAIR1M. Then, accurate fine-grained recognition becomes the key to achieve higher results. We compared with seven mainstream oriented object detection methods, including the single-stage and the two-stage methods. Table \ref{tab:mar20} shows the experimental results on MAR20 dataset. As can be seen, PETDet can also achieve state-of-the-art on the MAR20 dataset, surpassing Oriented R-CNN by 3.20 AP. In addition, PETDet significantly outperforms all other methods in $\mathrm{AP_{75}}$ and $\mathrm{AP_{50:95}}$, demonstrating that our PETDet can significantly improve the quality of localization.

\begin{table}[!t]
  \centering
  \caption{Comparison results on MAR20.}
  \begin{tblr}{
      stretch=1.05,
      column{2-4} = {c},
      cell{2}{1} = {c=4}{l},
      cell{6}{1} = {c=4}{l},
      hline{3} = {-}{},
      hline{6} = {1}{-}{},
      hline{7} = {-}{},
      vline{2-3} = {-}{},
    }
    \toprule
    Method                          & Backbone & $\mathrm{AP_{50:95}}$ & $\mathrm{AP_{50}}$ & $\mathrm{AP_{75}}$ \\
    \midrule
    \textit{One-stage methods:}                                                                                  \\
    RetinaNet \cite{retinanet}      & R-50-FPN & 43.78                 & 68.56              & 48.86              \\
    R$^3$Det \cite{r3det}           & R-50-FPN & 40.13                 & 61.18              & 48.13              \\
    S$^2$ANet \cite{s2anet}         & R-50-FPN & 41.82                 & 70.76              & 42.64              \\
    \textit{Two-stage methods:}                                                                                  \\
    Faster R-CNN \cite{faster-rcnn} & R-50-FPN & 47.57                 & 75.01              & 52.93              \\
    Gliding Vertex \cite{GV}        & R-50-FPN & 41.24                 & 80.28              & 37.60              \\
    RoI Transformer \cite{roitrans} & R-50-FPN & 56.43                 & 82.46              & 69.00              \\
    Oriented R-CNN \cite{orcnn}     & R-50-FPN & 58.14                 & 82.71              & 72.52              \\
    \textbf{PETDet (Ours)}          & R-50-FPN & \textbf{61.48}        & \textbf{85.91}     & \textbf{78.10}     \\
    \bottomrule
  \end{tblr}
  \label{tab:mar20}
\end{table}

\subsubsection{Quantitative Result on ShipRSImageNet}
We also conducted experiments on the ShipRSImageNet dataset, the results of which are shown in Table \ref{tab:shiprs}. Our proposed PETDet can also perform will on this dataset. Compared to Oriented R-CNN, PETDet introduces +3.14 $\mathrm{AP_{50}}$ and +3.79 $\mathrm{AP_{50:95}}$ gains, respectively. Both experiments on MAR20 and ShipRSImageNet show that without design for specific categories, PETDet also can achieve competitive performance in the single-class FGOD task.

\begin{table}[!t]
  \centering
  \caption{Comparison results on ShipRSImageNet.}
  \begin{tblr}{
      stretch=1.05,
      column{2-4} = {c},
      cell{2}{1} = {c=4}{l},
      cell{6}{1} = {c=4}{l},
      hline{3} = {-}{},
      hline{6} = {1}{-}{},
      hline{7} = {-}{},
      vline{2-3} = {-}{},
    }
    \toprule
    Method                          & Backbone & $\mathrm{AP_{50:95}}$ & $\mathrm{AP_{50}}$ & $\mathrm{AP_{75}}$ \\
    \midrule
    \textit{One-stage methods:}                                                                                  \\
    RetinaNet \cite{retinanet}      & R-50-FPN & 12.56                 & 20.08              & 14.14              \\
    R$^3$Det \cite{r3det}           & R-50-FPN & 12.82                 & 23.43              & 13.14              \\
    S$^2$ANet \cite{s2anet}         & R-50-FPN & 28.71                 & 49.39              & 30.43              \\
    \textit{Two-stage methods:}                                                                                  \\
    Faster R-CNN \cite{faster-rcnn} & R-50-FPN & 27.60                 & 54.75              & 24.09              \\
    Gliding Vertex \cite{GV}        & R-50-FPN & 28.67                 & 58.64              & 24.37              \\
    RoI Transformer \cite{roitrans} & R-50-FPN & 33.56                 & 60.98              & 32.17              \\
    Oriented R-CNN \cite{orcnn}     & R-50-FPN & 51.90                 & 71.76              & 63.69              \\
    \textbf{PETDet (Ours)}          & R-50-FPN & \textbf{55.69}        & \textbf{74.90}     & \textbf{67.86}     \\
    \bottomrule
  \end{tblr}
  \label{tab:shiprs}
\end{table}

\subsection{Ablation Study}
We conduct extensive experiments to verify the effectiveness of proposed module in our PETDet. We first ablate each component step by step to verify effectiveness of our QOPN, BCFN and ARL. Then, we respectively analyze specific effectiveness and optimal parameter settings of each component. Unless stated otherwise, all experiments for ablations are conducted on FAIR1M-v2.0 with a ResNet-50 backbone.

\subsubsection{Effectiveness of New Components}
We perform a componentwise ablation to thoroughly analyze the effect of the three key components in our PETDet on various FGOD datasets. As presented in Table \ref{tab:component}, all three components QOPN, BCFN and ARL are beneficial for performance on each dataset. Furthermore, it should be noted that three components give difference AP gains on difference datasets. On FAIR1M-v2.0 and MAR20, ARL contribute the most improvement (+2.42 $\mathrm{AP}$ and +2.79 $\mathrm{AP}$) by focusing on hard and important samples. On ShipRSImageNet, detectors will face numerous ships dense arranged with large aspect radios. With anchor-free paradigm and stronger architecture, QOPN can more obviously benefit it (+1.40 $\mathrm{AP}$) via better proposal generation.

\begin{table}[!tbp]
  \centering
  \caption{Effectiveness of Each Proposed Component.}
  \SetTblrInner{colsep=4pt}
  \begin{tblr}{
      cells = {c},
      cell{2}{1} = {r=4}{},
      cell{6}{1} = {r=4}{},
      cell{10}{1} = {r=4}{},
      vline{2} = {-}{},
      vline{5} = {-}{},
    }
    \toprule
    Dataset        & QOPN    & BCFN    & ARL     & $\mathrm{AP_{50:95}}$ & $\mathrm{AP_{50}}$ & $\mathrm{AP_{75}}$ \\
    \midrule
    FAIR1M-v2.0    &         &         &         & -                     & 44.26              & -                  \\
                   & $\surd$ &         &         & -                     & 45.00              & -                  \\
                   & $\surd$ & $\surd$ &         & -                     & 46.39              & -                  \\
                   & $\surd$ & $\surd$ & $\surd$ & -                     & \textbf{48.81}     & -                  \\
    \midrule
    MAR20          &         &         &         & 58.14                 & 82.71              & 72.52              \\
                   & $\surd$ &         &         & 58.46                 & 83.05              & 73.65              \\
                   & $\surd$ & $\surd$ &         & 58.71                 & 83.12              & 73.85              \\
                   & $\surd$ & $\surd$ & $\surd$ & \textbf{61.48}        & \textbf{85.91}     & \textbf{78.10}     \\
    \midrule
    ShipRSImageNet &         &         &         & 51.90                 & 71.76              & 63.69              \\
                   & $\surd$ &         &         & 52.81                 & 73.16              & 65.08              \\
                   & $\surd$ & $\surd$ &         & 54.88                 & 74.28              & 66.80              \\
                   & $\surd$ & $\surd$ & $\surd$ & \textbf{55.69}        & \textbf{74.90}     & \textbf{67.86}     \\
    \bottomrule
  \end{tblr}
  \label{tab:component}
\end{table}

\subsubsection{Ablation on QOPN}
In QOPN, LDAM is designed to decompose the BG/FG classification and regression, including the layer attention aggregation (LAA) and simple spatial attention (SSA). To thoroughly investigate the importance of LDAM, we perform a study of different network architectures of QOPN. The result is shown in Table \ref{tab:qopn_arch}, where stacked convs denotes that number of convolution layers in shared branch. We can observe that both LAA and SSA can bring improvements with negligible increment of FLOPs and model parameters. Benefited from LDAM, QOPN with 2 stacked convolutional layers can outperform than 4 stacked convolutional layers.

\begin{table}[!tbp]
  \centering
  \caption{Ablation Studies on different network architectures of QOPN.}
  \begin{tblr}{
      cells = {c},
      vline{4} = {-}{},
    }
    \toprule
    Stacked convs & LAA     & SSA     & $\mathrm{AP_{50}}$ & FLOPs(G) & \#params.(M) \\
    \midrule
    2             &         &         & 44.61              & 143.82   & 46.41        \\
    4             &         &         & 44.86              & 169.60   & 47.60        \\
    2             & $\surd$ &         & 44.84              & 143.85   & 46.81        \\
    2             & $\surd$ & $\surd$ & 45.00              & 143.86   & 46.81        \\
    \bottomrule
  \end{tblr}
  \label{tab:qopn_arch}
\end{table}

\par
To further substantiate the quality improvement of proposals generated by QOPN, we evaluate the performance of QOPN in terms of recall under different settings. Specifically, we respectively take 300, 500 and 1000 proposals, calculate the recall under the 0.5, 0.75 and 0.85 IoU thresholds. Then, the AR (the average recall within the range of 0.5 to 0.95 IoU thresholds) with different number of proposals will be calculated. We compare our QOPN with Oriented RPN (ORPN) proposed in Oriented R-CNN \cite{orcnn}. The experiments are conducted on the FAIR1M-v2.0 validation set, and the results are presented in Table \ref{tab:qopn_recall}. It can be found that our QOPN can not only significantly improve recall with 0.5 IoU threshold, also has significant gains at higher IoU threshold. This suggests that our QOPN can generate a greater number of well-localized proposals, which is crucial for subsequent fine-grained recognition in the second stage.

\begin{table}[!tbp]
  \centering
  \caption{Comparison of QOPN and ORPN in recall on FAIR1M-2.0 validation.}
  \begin{tblr}{
    cells = {c},
    cell{2}{1} = {r=2}{},
    cell{4}{1} = {r=2}{},
    cell{6}{1} = {r=2}{},
    vline{3} = {-}{},
    hline{4,6}={-}{},
      }
    \toprule
    \#proposals & Method & R50            & R75            & R85            & AR             \\
    \midrule
    300         & ORPN   & 0.663          & 0.317          & 0.103          & 0.343          \\
                & QOPN   & \textbf{0.907} & \textbf{0.515} & \textbf{0.169} & \textbf{0.513} \\
    500         & ORPN   & 0.718          & 0.336          & 0.103          & 0.368          \\
                & QOPN   & \textbf{0.927} & \textbf{0.524} & \textbf{0.171} & \textbf{0.523} \\
    1000        & ORPN   & 0.753          & 0.350          & 0.108          & 0.384          \\
                & QOPN   & \textbf{0.939} & \textbf{0.529} & \textbf{0.171} & \textbf{0.529} \\
    \bottomrule
  \end{tblr}
  \label{tab:qopn_recall}
\end{table}

\subsubsection{Ablation on BCFN}
As shown in Table IV, we explore different network designs in BCFN standing on the shoulders of QOPN. Here \textit{high-level} and \textit{low-level} refer to the direct utilization of $\left\{P_{3}, P_{4}, P_{5}, P_{6}, P_{7}\right\}$ or $\left\{P_{2}, P_{3}, P_{4}, P_{5}, P_{6} \right\}$ as the input of the second stage. And \textit{FPN-style} represents the FPN-like fusion method for comparison. In details, firstly the high-level feature map will be upsampled by nearest neighbors interpolation. Then, a 1 × 1 convolution is applied to adjust the number of channels. After that, an element-wise addition of high-level and low-level feature maps is performed, followed by a 3 × 3 convolution to mitigate the aliasing effect.

\par
Our results demonstrated that single-level feature is inadequate for FGOD tasks. In addition, low-level feature is more important for fine-grained recognition with richer local texture and patterns. Cross-level fusion is an effective way to enhance feature representation of proposals. Compared with FPN-style fusion, our BCFN can perform better cross-level fusion with lower FLOPs and parameters.

\begin{table}[!tbp]
  \centering
  \caption{Ablation Studies of different cross-level fusion approaches.}
  \begin{tblr}{
      cells = {c},
      vline{2} = {-}{},
    }
    \toprule
    Fusion method & $\mathrm{AP_{50}}$ & FLOPs(G) & \#params.(M) \\
    \midrule
    high-level    & 45.00              & 143.86   & 46.81        \\
    low-level     & 45.77              & 186.84   & 47.47        \\
    \midrule
    FPN-style     & 46.13              & 238.35   & 48.06        \\
    BCFN          & \textbf{46.39}     & 204.07   & 47.67        \\
    \bottomrule
  \end{tblr}
  \label{tab:ablation_bcfn}
\end{table}

\subsubsection{Ablation on ARL}

\begin{table}[!tbp]
  \centering
  \caption{Ablation Studies of NMS on proposals in recall on FAIR1M-2.0 validation.}
  \begin{tblr}{
    cells = {c},
    cell{2}{1} = {r=2}{},
    cell{4}{1} = {r=2}{},
    cell{6}{1} = {r=2}{},
    vline{3} = {-}{},
    hline{4,6} = {-}{},
      }
    \toprule
    \#proposals & $\mathrm{NMS_{proposal}}$ & R50            & R75            & R85            & AR             \\
    \midrule
    300         & $\surd$                   & \textbf{0.907} & 0.515          & 0.169          & 0.513          \\
                &                           & 0.865          & \textbf{0.549} & \textbf{0.225} & \textbf{0.521} \\
    500         & $\surd$                   & \textbf{0.927} & 0.524          & 0.171          & 0.523          \\
                &                           & 0.911          & \textbf{0.575} & \textbf{0.233} & \textbf{0.547} \\
    1000        & $\surd$                   & \textbf{0.939} & 0.529          & 0.171          & 0.529          \\
                &                           & 0.938          & \textbf{0.589} & \textbf{0.238} & \textbf{0.561} \\
    \bottomrule
  \end{tblr}
  \label{tab:nms_recall}
\end{table}

\par
When ARL is applied in the R-CNN head, the NMS on proposals is discarded in our configuration. Based on the proposals provided by QOPN, we first evaluate the recall rate under the conditions with and without NMS, as shown in Table \ref{tab:nms_recall}. On one hand, if NMS is not performed, R50 will decrease with fewer proposals since one ground truth will occupy more matched proposals. However, if we increase the number of proposals to 1000, this effect becomes negligible. On the other hand, the recall at higher IoU and the average recall benefit from discarding NMS on proposals. As we discussed, due to the inconsistency between localization and classification in the first-stage, high-scored proposals may not have the most accurate bounding boxes. Therefore, not performing the NMS on proposals improves the recall at high IoU thresholds.

\begin{table}[!tbp]
  \centering
  \caption{Comparison of cross-entropy loss (CE) and ARL with different post-processing settings of proposals.}
  \begin{tblr}{
      cells = {c},
      vline{4} = {-}{},
    }
    \toprule
    Loss & $\mathrm{NMS_{proposal}}$ & \#proposals & $\mathrm{AP_{50}}$ \\
    \midrule
    CE   & $\surd$                   & 2000        & 46.39              \\
    CE   &                           & 2000        & 46.70              \\
    CE   & $\surd$                   & 1000        & 46.22              \\
    CE   &                           & 1000        & 46.89              \\
    ARL  & $\surd$                   & 1000        & 47.05              \\
    ARL  &                           & 1000        & \textbf{48.81}     \\
    \bottomrule
  \end{tblr}
  \label{tab:arl}
\end{table}

\par
Table \ref{tab:arl} reports the comparison of cross-entropy loss and ARL under different post-processing settings of proposals. It can be found that reducing the number of proposals to 1000 will not significantly affect the performance. Besides, When using cross-entropy loss, an increase of 0.31 and 0.67 in AP can be achieved under 2000 and 1000 proposals by removing NMS on proposals, respectively. These results meet our expectations, as 1000 proposals are already sufficient to achieve a high recall rate, and more high-quality proposals can be reserved without NMS on proposals. After applying proposed ARL, 47.05 AP can be achieved with 1000 proposals after NMS, surpassing cross-entropy loss of 0.83 AP under the same settings. Furthermore, if NMS is not applied, we can get a larger extra gain of 1.76 AP with ARL. This suggests that ARL can better utilize noisy proposals to promote fine-grained recognition learning.

\begin{table}[!tbp]
  \centering
  \caption{Detection performances by setting different values of $\gamma$ and $\beta$ in ARL.}
  \begin{tblr}{
      cells = {c},
      vline{2} = {-}{},
    }
    \toprule
    \diagbox{$\gamma$}{$\beta$} & 1.0   & 1.5   & 2.0   & 2.5            & 3.0   \\
    \hline
    1.0                         & 47.23 & 48.08 & 48.55 & 48.73          & 48.36 \\
    1.5                         & 47.40 & 48.07 & 48.58 & \textbf{48.81} & 48.47 \\
    2.0                         & 48.45 & 48.09 & 48.51 & 48.57          & 48.53 \\
    \bottomrule
  \end{tblr}
  \label{tab:arl_hyper}
\end{table}

\par
We also conduct experiments to investigate the robustness of two hyperparameters $\gamma$ and $\beta$ set in ARL. Here, $\gamma$ can reduce the weight of easy negative samples, while $\beta$ is to control the relative scales between different positive sample by the quality of proposals. We make a grid search to investigate the impact of the hyperparameters, as shown in Table \ref{tab:arl_hyper}. It can be seen that both $\gamma$ and $\beta$ set small will decrease the performance. When $\gamma$ ranges from 1.5 to 2.0 and $\beta$ ranges from 1.5 to 3.0, the performance is no longer sensitive to these two hyperparameters. We adopt $\gamma=1.5$ and $\beta=2.5$ as default in all other experiments. With this combination, our PETDet can achieve 48.81 AP on FAIR1M-v2.0 dataset. Notice that the default number of proposals and hyperparameters selection is set slightly different on other dataset. For instance, The detection task in the MAR20 dataset is much easier, where a less number of proposals can ensure the recall rate. Consequently, we decrease the number of proposals taken into the second stage, and set $\beta$ in ARL lower to balance the weight between positive and negative samples.

\section{Conclusion} \label{sec:conclusion}
In this paper, we explore two-stage FGOD methods from a novel perspective. We firstly highlight the asset of region proposal in FGOD from the perspective of multi-task learning. Therefore, we present an improved two-stage FGOD method with proposal enhancement called PETDet. Our model mainly contributes in three aspects: 1) improves the proposal quality (QOPN module); 2) leverages cross-level discriminative features (BCFN module); 3) re-weights proposals to focus on high-quality samples (ARL module). Extensive experimental results on four universal datasets with all-sided ablation studies on each module demonstrate the effectiveness of PETDet. Compared to other methods, PETDet not only reaches state-of-the-art performance but also achieves favorable accuracy-versus-speed trade-off. Although PETDet shows great performance, the design is only suitable for two-stage pipeline and only evaluated on large-scale datasets. Our future work will extend our model from two main aspects: 1) more architecture-efficient from including transformer-based methods; 2) few-shot FGOD adaptation to meet more practical data-scarcity applications. We hope that our exploration of  multi-task interaction in PETDet can contribute to further advancements in the field of FGOD.

\ifCLASSOPTIONcaptionsoff
  \newpage
\fi
\printbibliography

@article{fair1m,
  title   = {{FAIR1M}: A Benchmark Dataset for Fine-grained Object Recognition in High-resolution Remote Sensing Imagery},
  author  = {Sun, Xian and Wang, Peijin and Yan, Zhiyuan and Xu, Feng and Wang, Ruiping and Diao, Wenhui and Chen, Jin and Li, Jihao and Feng, Yingchao and Xu, Tao and others},
  journal = {ISPRS J. Photogramm. Remote Sens.},
  volume  = {184},
  pages   = {116--130},
  month   = {2},
  year    = {2022}
}

@article{mar20,
  title   = {{MAR20}: A Benchmark for Military Aircraft Recognition in Remote Sensing Images},
  author  = {Yu,WenQi and Cheng,Gong and Wang,MeiJun and Yao,YanQing and Xie,XingXing and Yao,XiWen and Han,JunWei},
  year    = {2022},
  journal = {National Remote Sensing Bulletin}
}

@article{shiprs,
  title   = {{ShipRSImageNet}: A large-scale fine-grained dataset for ship detection in high-resolution optical remote sensing images},
  author  = {Zhang, Zhengning and Zhang, Lin and Wang, Yue and Feng, Pengming and He, Ran},
  journal = {IEEE J. Sel. Topics Appl. Earth Observ. Remote Sens},
  volume  = {14},
  pages   = {8458--8472},
  year    = {2021}
}

@inproceedings{orcnn,
  title     = {{Oriented R-CNN} for Object Detection},
  author    = {Xie, Xingxing and Cheng, Gong and Wang, Jiabao and Yao, Xiwen and Han, Junwei},
  booktitle = {Proc. IEEE/CVF Int. Conf. Comput. Vis. (ICCV)},
  pages     = {3520--3529},
  month     = {10},
  year      = {2021}
}

@inproceedings{roitrans,
  title     = {Learning Roi Transformer for Oriented Object Detection in Aerial Images},
  author    = {Ding, Jian and Xue, Nan and Long, Yang and Xia, Gui-Song and Lu, Qikai},
  booktitle = {Proc. IEEE/CVF Conf. Comput. Vis. Pattern Recognit. (CVPR)},
  pages     = {2849--2858},
  month     = {6},
  year      = {2019}
}

@article{tioe,
  title   = {Task interleaving and orientation estimation for high-precision oriented object detection in aerial images},
  author  = {Qi Ming and Lingjuan Miao and Zhiqiang Zhou and Junjie Song and Yunpeng Dong and Xue Yang},
  journal = {ISPRS J. Photogramm. Remote Sens.},
  volume  = {196},
  pages   = {241-255},
  month   = {2},
  year    = {2023}
}

@article{ridet,
  title   = {Optimization for Arbitrary-Oriented Object Detection via Representation Invariance Loss},
  author  = {Ming, Qi and Miao, Lingjuan and Zhou, Zhiqiang and Yang, Xue and Dong, Yunpeng},
  journal = {IEEE Geosci. Remote Sens. Lett.},
  volume  = {19},
  month   = {10},
  year    = {2022},
  note    = {{A}rt. no. 8021505.}
}

@inproceedings{scrdet,
  title     = {{SCRDet}: Towards More Robust Detection for Small, Cluttered and Rotated Objects},
  author    = {Yang, Xue and Yang, Jirui and Yan, Junchi and Zhang, Yue and Zhang, Tengfei and Guo, Zhi and Sun, Xian and Fu, Kun},
  booktitle = {Proc. IEEE Int. Conf. Comput. Vis. (ICCV)},
  pages     = {8232--8241},
  month     = {10},
  year      = {2019}
}

@inproceedings{r3det,
  title     = {{R3Det}: Refined Single-stage Detector with Feature Refinement for Rotating Object},
  author    = {Yang, Xue and Yan, Junchi and Feng, Ziming and He, Tao},
  booktitle = {Proc. AAAI Conf. Artif. Intell.},
  pages     = {3163--3171},
  year      = {2021}
}

@article{s2anet,
  title   = {Align Deep Features for Oriented Object Detection},
  author  = {Han, Jiaming and Ding, Jian and Li, Jie and Xia, Gui-Song},
  journal = {IEEE Trans. Geosci. Remote Sens.},
  volume  = {60},
  month   = {3},
  year    = {2021},
  note    = {{A}rt. no. 5602511}
}

@inproceedings{doublehead,
  title     = {Rethinking Classification and Localization for Object Detection},
  author    = {Wu, Yue and Chen, Yinpeng and Yuan, Lu and Liu, Zicheng and Wang, Lijuan and Li, Hongzhi and Fu, Yun},
  booktitle = {Proc. IEEE/CVF Conf. Comput. Vis. Pattern Recognit. (CVPR)},
  pages     = {10186--10195},
  month     = {6},
  year      = {2020}
}

@article{rrpn,
  title   = {Arbitrary-oriented Scene Text Detection via Rotation Proposals},
  author  = {Ma, Jianqi and Shao, Weiyuan and Ye, Hao and Wang, Li and Wang, Hong and Zheng, Yingbin and Xue, Xiangyang},
  journal = {IEEE Trans. Multimedia},
  volume  = {20},
  number  = {11},
  pages   = {3111--3122},
  month   = {3},
  year    = {2018}
}

@inproceedings{redet,
  author    = {Han, Jiaming and Ding, Jian and Xue, Nan and Xia, Gui-Song},
  title     = {{ReDet}: A Rotation-equivariant Detector for Aerial Object Detection},
  booktitle = {Proc. IEEE/CVF Conf. Comput. Vis. Pattern Recognit. (CVPR)},
  month     = {6},
  year      = {2021},
  pages     = {2786-2795}
}

@article{aopg,
  title   = {Object Detection in Optical Remote Sensing Images: A Survey and A New Benchmark},
  author  = {Li, Ke and Wan, Gang and Cheng, Gong and Meng, Liqiu and Han, Junwei},
  journal = {ISPRS J. Photogramm. Remote Sens.},
  volume  = {159},
  pages   = {296--307},
  month   = {1},
  year    = {2020}
}

@inproceedings{gwd,
  title     = {Rethinking Rotated Object Detection with Gaussian Wasserstein Distance Loss},
  author    = {Yang, Xue and Yan, Junchi and Qi, Ming and Wang, Wentao and Xiaopeng, Zhang and Qi, Tian},
  booktitle = {IEEE Int. Conf. Mach. Learn.},
  pages     = {11830--11841},
  year      = {2021}
}

@inproceedings{kld,
  title     = {Learning High-Precision Bounding Box for Rotated Object Detection via Kullback-Leibler Divergence},
  author    = {Yang, Xue and Yang, Xiaojiang and Yang, Jirui and Ming, Qi and Wang, Wentao and Tian, Qi and Yan, Junchi},
  booktitle = {Proc. Conf. Adv. Neural Inf. Process. Syst.},
  pages     = {18381--18394},
  year      = {2021}
}

@inproceedings{psc,
  author    = {Yu, Yi and Da, Feipeng},
  title     = {Phase-Shifting Coder: Predicting Accurate Orientation in Oriented Object Detection},
  booktitle = {Proc. IEEE/CVF Conf. Comput. Vis. Pattern Recognit. (CVPR)},
  pages     = {13354-13363},
  month     = {6},
  year      = {2023}
}

@inproceedings{dal,
  title     = {Dynamic Anchor Learning for Arbitrary-Oriented Object Detection},
  author    = {Ming, Qi and Zhou, Zhiqiang and Miao, Lingjuan and Zhang, Hongwei and Li, Linhao},
  booktitle = {Proc. AAAI Conf. Artif. Intell},
  pages     = {2355--2363},
  year      = {2021}
}

@article{cfc,
  title   = {{CFC-Net}: A Critical Feature Capturing Network for Arbitrary-Oriented Object Detection in Remote-Sensing Images},
  author  = {Ming, Qi and Miao, Lingjuan and Zhou, Zhiqiang and Dong, Yunpeng},
  journal = {IEEE Trans. Geosci. Remote Sens.},
  volume  = {60},
  pages   = {1-14},
  year    = {2022}
}

@article{sfr,
  title   = {{SFRNet}: Fine-Grained Oriented Object Recognition via Separate Feature Refinement},
  author  = {Cheng, Gong and Li, Qingyang and Wang, Guangxing and Xie, Xingxing and Min, Lingtong and Han, Junwei},
  journal = {IEEE Trans. Geosci. Remote Sens.},
  volume  = {61},
  year    = {2023},
  note    = {{A}rt no. 5610510}
}

@inproceedings{part,
  title     = {Part-based R-CNNs for Fine-grained Category Detection},
  author    = {Zhang, Ning and Donahue, Jeff and Girshick, Ross and Darrell, Trevor},
  booktitle = {Proc. Eur. Conf. Comput. Vis. (ECCV)},
  pages     = {834--849},
  year      = {2014}
}

@inproceedings{racnn,
  title     = {Look Closer to See Better: Recurrent Attention Convolutional Neural Network for Fine-grained Image Recognition},
  author    = {Fu, Jianlong and Zheng, Heliang and Mei, Tao},
  booktitle = {Proc. IEEE/CVF Conf. Comput. Vis. Pattern Recognit. (CVPR)},
  pages     = {4438--4446},
  year      = {2017},
  month     = {7}
}

@inproceedings{bilinear,
  title     = {Bilinear CNN Models for Fine-grained Visual Recognition},
  author    = {Lin, TsungYu and RoyChowdhury, Aruni and Maji, Subhransu},
  booktitle = {Proc. IEEE/CVF Int. Conf. Comput. Vis. (ICCV)},
  pages     = {1449--1457},
  year      = {2015},
  month     = {12}
}

@inproceedings{querydet,
  title     = {{QueryDet}: Cascaded Sparse Query for Accelerating High-resolution Small Object Detection},
  author    = {Yang, Chenhongyi and Huang, Zehao and Wang, Naiyan},
  booktitle = {Proc. IEEE/CVF Conf. Comput. Vis. Pattern Recognit. (CVPR)},
  pages     = {13668--13677},
  year      = {2022},
  month     = {6}
}

@inproceedings{fcos,
  title     = {{FCOS}: Fully Convolutional One-stage Object Detection},
  author    = {Tian, Zhi and Shen, Chunhua and Chen, Hao and He, Tong},
  booktitle = {Proc. IEEE/CVF Int. Conf. Comput. Vis. (ICCV)},
  pages     = {9627--9636},
  month     = {10},
  year      = {2019}
}

@inproceedings{atss,
  title     = {Bridging the Gap between Anchor-based and Anchor-free Detection via Adaptive Training Sample Selection},
  author    = {Zhang, Shifeng and Chi, Cheng and Yao, Yongqiang and Lei, Zhen and Li, Stan Z},
  booktitle = {Proc. IEEE/CVF Conf. Comput. Vis. Pattern Recognit. (CVPR)},
  pages     = {9759--9768},
  month     = {6},
  year      = {2020}
}

@inproceedings{retinanet,
  title     = {Focal Loss for Dense Object Detection},
  author    = {Lin, Tsung-Yi and Goyal, Priya and Girshick, Ross and He, Kaiming and Doll{\'a}r, Piotr},
  booktitle = {Proc. IEEE Int. Conf. Comput. Vis. (ICCV)},
  pages     = {2980--2988},
  month     = {10},
  year      = {2017}
}

@article{centernet2,
  title   = {Probabilistic Two-stage Detection},
  author  = {Zhou, Xingyi and Koltun, Vladlen and Kr{\"a}henb{\"u}hl, Philipp},
  journal = {arXiv:2103.07461},
  year    = {2021}
}

@article{GV,
  title   = {Gliding Vertex on the Horizontal Bounding Box for Multi-oriented Object Detection},
  author  = {Xu, Yongchao and Fu, Mingtao and Wang, Qimeng and Wang, Yukang and Chen, Kai and Xia, Gui-Song and Bai, Xiang},
  journal = {IEEE Trans. Pattern Anal. Mach. Intell.},
  volume  = {43},
  number  = {4},
  pages   = {1452--1459},
  month   = {4},
  year    = {2020}
}

@inproceedings{fpn,
  title     = {Feature Pyramid Networks for Object Detection},
  author    = {Lin, TsungYi and Doll{\'a}r, Piotr and Girshick, Ross and He, Kaiming and Hariharan, Bharath and Belongie, Serge},
  booktitle = {Proc. IEEE Conf. Comput. Vis. Pattern Recognit. (CVPR)},
  pages     = {2117--2125},
  year      = {2017},
  month     = {7}
}

@inproceedings{resnet,
  title     = {Deep Residual Learning for Image Recognition},
  author    = {He, Kaiming and Zhang, Xiangyu and Ren, Shaoqing and Sun, Jian},
  booktitle = {Proc. IEEE Conf. Comput. Vis. Pattern Recognit. (CVPR)},
  pages     = {770--778},
  month     = {6},
  year      = {2016}
}

@inproceedings{mmrotate,
  title     = {{MMRotate}: A Rotated Object Detection Benchmark using PyTorch},
  author    = {Zhou, Yue and Yang, Xue and Zhang, Gefan and Wang, Jiabao and Liu, Yanyi and
               Hou, Liping and Jiang, Xue and Liu, Xingzhao and Yan, Junchi and Lyu, Chengqi and
               Zhang, Wenwei and Chen, Kai},
  booktitle = {Proc. ACM Int. Conf. Multimedia},
  year      = {2022}
}

@inproceedings{macnn,
  title     = {Learning Multi-attention Convolutional Neural Network for Fine-grained Image Recognition},
  author    = {Zheng, Heliang and Fu, Jianlong and Mei, Tao and Luo, Jiebo},
  booktitle = {Proc. IEEE Int. Conf. Comput. Vis. (ICCV)},
  pages     = {5209--5217},
  year      = {2017},
  month     = {10}
}

@article{mcloss,
  title   = {The Devil is in the Channels: Mutual-channel Loss for Fine-grained Image Classification},
  author  = {Chang, Dongliang and Ding, Yifeng and Xie, Jiyang and Bhunia, Ayan Kumar and Li, Xiaoxu and Ma, Zhanyu and Wu, Ming and Guo, Jun and Song, Yi-Zhe},
  journal = {IEEE Trans. Image Process.},
  volume  = {29},
  pages   = {4683--4695},
  year    = {2020}
}

@article{gicnet,
  title   = {Classification Matters More: Global Instance Contrast for Fine-Grained SAR Aircraft Detection},
  author  = {Zhao, Danpei and Chen, Ziqiang and Gao, Yue and Shi, Zhenwei},
  journal = {IEEE Trans. Geosci. Remote Sens.},
  volume  = {61},
  month   = {3},
  year    = {2023},
  note    = {{A}rt no. 5203815}
}

@article{osfd,
  title   = {Multigranularity Self-Attention Network for Fine-Grained Ship Detection in Remote Sensing Images},
  author  = {Ouyang, Lihan and Fang, Leyuan and Ji, Xinyu},
  journal = {IEEE J. Sel. Topics Appl. Earth Observ. Remote Sens.},
  volume  = {15},
  pages   = {9722--9732},
  year    = {2022}
}

@article{iscl,
  title   = {Instance Switching-Based Contrastive Learning for Fine-Grained Airplane Detection},
  author  = {Zeng, Lanxin and Guo, Haowen and Yang, Wen and Yu, Huai and Yu, Lei and Zhang, Peng and Zou, Tongyuan},
  journal = {IEEE Trans. Geosci. Remote Sens.},
  volume  = {60},
  year    = {2022},
  note    = {{A}rt no. 5633416}
}

@article{fgfe,
  author  = {Zhou, Yong and Wang, Sifan and Zhao, Jiaqi and Zhu, Hancheng and Yao, Rui},
  journal = {IEEE Geosci. Remote Sens. Lett.},
  title   = {Fine-Grained Feature Enhancement for Object Detection in Remote Sensing Images},
  year    = {2022},
  volume  = {19},
  note    = {{A}rt no. 6508305}
}

@article{acd,
  author  = {Jia, Hecheng and Guo, Qian and Chen, Jin and Wang, Feng and Wang, Haipeng and Xu, Feng},
  journal = {IEEE J. Sel. Topics Appl. Earth Observ. Remote Sens.},
  title   = {Adaptive Component Discrimination Network for Airplane Detection in Remote Sensing Images},
  year    = {2021},
  volume  = {14},
  pages   = {7699-7713}
}

@article{faster-rcnn,
  author  = {Ren, Shaoqing and He, Kaiming and Girshick, Ross and Sun, Jian},
  journal = {IEEE Trans. Pattern Anal. Mach. Intell.},
  title   = {{Faster R-CNN}: Towards Real-Time Object Detection with Region Proposal Networks},
  year    = {2017},
  volume  = {39},
  number  = {6},
  pages   = {1137-1149}
}

@inproceedings{cascade-rcnn,
  title     = {{Cascade R-CNN: D}elving into High Quality Object Detection},
  author    = {Cai, Zhaowei and Vasconcelos, Nuno},
  booktitle = {Proc. IEEE/CVF Conf. Comput. Vis. Pattern Recognit. (CVPR)},
  pages     = {6154--6162},
  year      = {2018},
  month     = {6}
}

@inproceedings{libra-rcnn,
  title     = {{Libra R-CNN: T}owards Balanced Learning for Object Detection},
  author    = {Pang, Jiangmiao and Chen, Kai and Shi, Jianping and Feng, Huajun and Ouyang, Wanli and Dahua Lin},
  booktitle = {Proc. IEEE/CVF Conf. Comput. Vis. Pattern Recognit. (CVPR)},
  pages     = {821--830},
  year      = {2019},
  month     = {6}
}

@inproceedings{cornernet,
  title     = {{CornerNet}: Detecting objects as paired keypoints},
  author    = {Law, Hei and Deng, Jia},
  booktitle = {Proc. Eur. Conf. Comput. Vis. (ECCV)},
  pages     = {734--750},
  year      = {2018}
}

@inproceedings{centernet,
  title     = {{CenterNet}: Keypoint triplets for object detection},
  author    = {Duan, Kaiwen and Bai, Song and Xie, Lingxi and Qi, Honggang and Huang, Qingming and Tian, Qi},
  booktitle = {Proc. IEEE Int. Conf. Comput. Vis. (ICCV)},
  pages     = {6569--6578},
  year      = {2019},
  month     = {10}
}

@article{gfl,
  title   = {Generalized focal loss: Learning qualified and distributed bounding boxes for dense object detection},
  author  = {Li, Xiang and Wang, Wenhai and Wu, Lijun and Chen, Shuo and Hu, Xiaolin and Li, Jun and Tang, Jinhui and Yang, Jian},
  journal = {Proc. Adv. Neural Inf. Process. Syst.},
  volume  = {33},
  pages   = {21002--21012},
  year    = {2020}
}

@article{voc,
  title   = {The pascal visual object classes (voc) challenge},
  author  = {Everingham, Mark and Van Gool, Luc and Williams, Christopher KI and Winn, John and Zisserman, Andrew},
  journal = {Int. J. Comput. Vis.},
  volume  = {88},
  pages   = {303--338},
  year    = {2010}
}

@article{foveabox,
  title   = {{Foveabox}: Beyound anchor-based object detection},
  author  = {Kong, Tao and Sun, Fuchun and Liu, Huaping and Jiang, Yuning and Li, Lei and Shi, Jianbo},
  journal = {IEEE Trans. Image Process.},
  volume  = {29},
  pages   = {7389--7398},
  year    = {2020}
}

@inproceedings{paa,
  title     = {Probabilistic anchor assignment with iou prediction for object detection},
  author    = {Kim, Kang and Lee, Hee Seok},
  booktitle = {Proc. Eur. Conf. Comput. Vis. (ECCV)},
  pages     = {355--371},
  year      = {2020}
}

@inproceedings{ota,
  title     = {{OTA}: Optimal transport assignment for object detection},
  author    = {Ge, Zheng and Liu, Songtao and Li, Zeming and Yoshie, Osamu and Sun, Jian},
  booktitle = {Proc. IEEE/CVF Conf. Comput. Vis. Pattern Recognit. (CVPR)},
  pages     = {303--312},
  year      = {2021},
  month     = {6}
}

@inproceedings{vfnet,
  title     = {{VarifocalNet}: An iou-aware dense object detector},
  author    = {Zhang, Haoyang and Wang, Ying and Dayoub, Feras and Sunderhauf, Niko},
  booktitle = {Proc. IEEE/CVF Conf. Comput. Vis. Pattern Recognit. (CVPR)},
  pages     = {8514--8523},
  year      = {2021},
  month     = {6}
}

@inproceedings{dw,
  title     = {A dual weighting label assignment scheme for object detection},
  author    = {Li, Shuai and He, Chenhang and Li, Ruihuang and Zhang, Lei},
  booktitle = {Proc. IEEE/CVF Conf. Comput. Vis. Pattern Recognit. (CVPR)},
  pages     = {9387--9396},
  year      = {2022},
  month     = {6}
}

@inproceedings{r2cnn,
  title     = {{R2CNN}: Rotational region cnn for arbitrarily-oriented scene text detection},
  author    = {Jiang, Yingying and Zhu, Xiangyu and Wang, Xiaobing and Yang, Shuli and Li, Wei and Wang, Hua and Fu, Pei and Luo, Zhenbo},
  booktitle = {Int. Conf. Pattern Recognit. (ICPR)},
  pages     = {3610--3615},
  year      = {2018}
}

@article{orienteddetr,
  title   = {Oriented object detection with transformer},
  author  = {Ma, Teli and Mao, Mingyuan and Zheng, Honghui and Gao, Peng and Wang, Xiaodi and Han, Shumin and Ding, Errui and Zhang, Baochang and Doermann, David},
  journal = {arXiv:2106.03146},
  year    = {2021}
}

@article{ao2,
  title   = {{AO2-DETR}: Arbitrary-oriented object detection transformer},
  author  = {Dai, Linhui and Liu, Hong and Tang, Hao and Wu, Zhiwei and Song, Pinhao},
  journal = {IEEE Trans. Circuits Syst. Video Technol.},
  year    = {2022},
  volume  = {33},
  pages   = {2342--2356},
  year    = {2018},
  month   = {5}
}

@article{ars,
  title   = {{ARS-DETR}: Aspect ratio sensitive oriented object detection with transformer},
  author  = {Zeng, Ying and Yang, Xue and Li, Qingyun and Chen, Yushi and Yan, Junchi},
  journal = {arXiv:2303.04989},
  year    = {2023}
}

@inproceedings{d2q,
  title     = {{D2Q-DETR}: Decoupling and Dynamic Queries for Oriented Object Detection with Transformers},
  author    = {Zhou, Qiang and Yu, Chaohui and Wang, Zhibin and Wang, Fan},
  booktitle = {Proc. IEEE Int. Conf. Acoust. Speech Signal Process. (ICASSP)},
  year      = {2023}
}

@article{rhino,
  title   = {{RHINO}: Rotated DETR with Dynamic Denoising via Hungarian Matching for Oriented Object Detection},
  author  = {Lee, Hakjin and Song, Minki and Koo, Jamyoung and Seo, Junghoon},
  journal = {arXiv:2305.07598},
  year    = {2023}
}

@inproceedings{layerscale,
  title     = {Going deeper with image transformers},
  author    = {Touvron, Hugo and Cord, Matthieu and Sablayrolles, Alexandre and Synnaeve, Gabriel and J{\'e}gou, Herv{\'e}},
  booktitle = {Proc. IEEE Int. Conf. Comput. Vis. (ICCV)},
  pages     = {32--42},
  year      = {2021},
  month     = {10}
}

@inproceedings{visualizing,
  title     = {Visualizing and understanding convolutional networks},
  author    = {Zeiler, Matthew D and Fergus, Rob},
  booktitle = {Proc. Eur. Conf. Comput. Vis. (ECCV)},
  pages     = {818--833},
  year      = {2014}
}

@article{mlb,
  title   = {Hadamard product for low-rank bilinear pooling},
  author  = {Kim, Jin-Hwa and On, Kyoung-Woon and Lim, Woosang and Kim, Jeonghee and Ha, Jung-Woo and Zhang, Byoung-Tak},
  journal = {arXiv:1610.04325},
  year    = {2016}
}

@inproceedings{glu,
  title     = {Three new graphical models for statistical language modelling},
  author    = {Mnih, Andriy and Hinton, Geoffrey},
  booktitle = {Proc. Int. Conf. Mach. Learn.},
  pages     = {641--648},
  year      = {2007}
}

@article{gluvariants,
  title   = {{GLU} variants improve transformer},
  author  = {Shazeer, Noam},
  journal = {arXiv:2002.05202},
  year    = {2020}
}

@article{kd,
  title   = {Efficient Fine-Grained Object Recognition in High-Resolution Remote Sensing Images From Knowledge Distillation to Filter Grafting},
  author  = {Wang, Liuqian and Zhang, Jing and Tian, Jimiao and Li, Jiafeng and Zhuo, Li and Tian, Qi},
  journal = {IEEE Trans. Geosci. Remote Sens.},
  volume  = {61},
  year    = {2023},
  note    = {{A}rt no. 4701016}
}

@inproceedings{tood,
  title     = {{TOOD}: Task-aligned one-stage object detection},
  author    = {Feng, Chengjian and Zhong, Yujie and Gao, Yu and Scott, Matthew R and Huang, Weilin},
  booktitle = {Proc. IEEE Int. Conf. Comput. Vis. (ICCV)},
  pages     = {3490--3499},
  year      = {2021},
  month     = {10}
}

@inproceedings{oreppoints,
  title     = {Oriented reppoints for aerial object detection},
  author    = {Li, Wentong and Chen, Yijie and Hu, Kaixuan and Zhu, Jianke},
  booktitle = {Proc. IEEE/CVF Conf. Comput. Vis. Pattern Recognit. (CVPR)},
  pages     = {1829--1838},
  year      = {2022},
  month     = {6}
}

@article{dodet,
  title   = {Dual-aligned oriented detector},
  author  = {Cheng, Gong and Yao, Yanqing and Li, Shengyang and Li, Ke and Xie, Xingxing and Wang, Jiabao and Yao, Xiwen and Han, Junwei},
  journal = {IEEE Trans. Geosci. Remote Sens.},
  volume  = {60},
  year    = {2022},
  note    = {{A}rt no. 5618111}
}

@inproceedings{dynamic,
  title     = {{Dynamic R-CNN}: Towards high quality object detection via dynamic training},
  author    = {Zhang, Hongkai and Chang, Hong and Ma, Bingpeng and Wang, Naiyan and Chen, Xilin},
  booktitle = {Proc. Eur. Conf. Comput. Vis. (ECCV)},
  pages     = {260--275},
  year      = {2020}
}

@article{sis,
  title   = {Fine-Grained Ship Detection in High-Resolution Satellite Images With Shape-Aware Feature Learning},
  author  = {Guo, Bo and Zhang, Ruixiang and Guo, Haowen and Yang, Wen and Yu, Huai and Zhang, Peng and Zou, Tongyuan},
  journal = {IEEE J. Sel. Topics Appl. Earth Observ. Remote Sens.},
  volume  = {16},
  pages   = {1914--1926},
  year    = {2023}
}

@article{dosr,
  title   = {Fine-grained recognition for oriented ship against complex scenes in optical remote sensing images},
  author  = {Han, Yaqi and Yang, Xinyi and Pu, Tian and Peng, Zhenming},
  journal = {IEEE Trans. Geosci. Remote Sens.},
  volume  = {60},
  pages   = {1--18},
  year    = {2022},
  note    = {{A}rt no. 5612318}
}

@article{pcl,
  title   = {PCLDet: Prototypical Contrastive Learning for Fine-Grained Object Detection in Remote Sensing Images},
  author  = {Ouyang, Lihan and Guo, Guangmiao and Fang, Leyuan and Ghamisi, Pedram and Yue, Jun},
  journal = {IEEE Trans. Geosci. Remote Sens.},
  volume  = {61},
  pages   = {1-11},
  year    = {2023},
  note    = {{A}rt no. 5613911}
}

@article{ubc,
  author  = {Huang, Xingliang and Chen, Kaiqiang and Tang, Deke and Liu, Chenglong and Ren, Libo and Sun, Zheng and Hänsch, Ronny and Schmitt, Michael and Sun, Xian and Huang, Hai and Mayer, Helmut},
  journal = {IEEE Trans. Geosci. Remote Sens.},
  title   = {Urban Building Classification {(UBC)} V2—A Benchmark for Global Building Detection and Fine-Grained Classification From Satellite Imagery},
  year    = {2023},
  volume  = {61},
  number  = {},
  pages   = {1-16},
  note    = {{A}rt no. 5620116}
}

@article{low,
  title   = {A low-cost UAV framework towards ornamental plant detection and counting in the wild},
  author  = {Bayraktar, Ertugrul and Basarkan, Muhammed Enes and Celebi, Numan},
  journal = {ISPRS J. Photogramm. Remote Sens.},
  volume  = {167},
  pages   = {1--11},
  year    = {2020},
  month   = {9}
}

@article{traffic,
  title   = {Traffic congestion-aware graph-based vehicle rerouting framework from aerial imagery},
  journal = {Eng. Appl. Artif. Intell.},
  volume  = {119},
  pages   = {105769},
  year    = {2023},
  issn    = {0952-1976},
  author  = {Ertugrul Bayraktar and Burla Nur Korkmaz and Aras Umut Erarslan and Numan Celebi}
}

@inproceedings{dota,
  title     = {{DOTA}: A large-scale dataset for object detection in aerial images},
  author    = {Xia, Gui-Song and Bai, Xiang and Ding, Jian and Zhu, Zhen and Belongie, Serge and Luo, Jiebo and Datcu, Mihai and Pelillo, Marcello and Zhang, Liangpei},
  booktitle = {Proc. IEEE/CVF Conf. Comput. Vis. Pattern Recognit. (CVPR)},
  pages     = {3974--3983},
  year      = {2018},
  month     = {6}
}

@article{unsupervised,
  title     = {Unsupervised Domain Adaptation with Debiased Contrastive Learning and Support-Set Guided Pseudo Labeling for Remote Sensing Images},
  author    = {Biswas, Debojyoti and Tesic, Jelena},
  year      = {2023},
  publisher = {TechRxiv}
}

@article{fdd,
  author  = {Wang, Chenxu and Zhao, Danpei and Qi, Xinhu and Liu, Zhuoran and Shi, Zhenwei},
  journal = {IEEE Trans. Geosci. Remote Sens.},
  title   = {A Hierarchical Decoder Architecture for Multilevel Fine-Grained Disaster Detection},
  year    = {2023},
  volume  = {61},
  number  = {},
  pages   = {1-14},
  note    = {{A}rt no. 5607114}
}

\end{document}